%% file: main.tex
\documentclass[10pt,twocolumn,letterpaper]{article}

\PassOptionsToPackage{table,xcdraw}{xcolor}
\usepackage[pagenumbers]{cvpr} 

\usepackage{listings}
\usepackage{afterpage}

\input{preamble}
\definecolor{cvprblue}{rgb}{0.21,0.49,0.74}
\usepackage[pagebackref,breaklinks,colorlinks,allcolors=cvprblue]{hyperref}
\usepackage{multirow}
\usepackage{amssymb}
\usepackage{pifont}
\usepackage[most]{tcolorbox} 
\newcommand{\cmark}{\checkmark}     
\newcommand{\xmark}{\ding{55}}       
\usepackage{adjustbox}
\def\paperID{*****} 
\def\confName{CVPR}
\def\confYear{2026}

\title{\textit{ChartNet}: A Million-Scale, High-Quality Multimodal Dataset\\ for Robust Chart Understanding}

\author{%
  \parbox{\textwidth}{\centering
    Jovana Kondic$^{1}$\thanks{Correspondence: \tt jkondic@mit.edu},
    Pengyuan Li$^{3}$,
    Dhiraj Joshi$^{3}$,
    Isaac Sanchez$^{2}$,
    Ben Wiesel$^{3}$,
    Shafiq Abedin$^{3}$,\\
    Amit Alfassy$^{3}$,
    Eli Schwartz$^{3}$,
    Daniel Caraballo$^{3}$,
    Yagmur Gizem Cinar$^{3}$,
    Florian Scheidegger$^{3}$,\\
    Steven I.\ Ross$^{3}$,
    Daniel Karl I.\ Weidele$^{2}$,
    Hang Hua$^{2}$,
    Ekaterina Arutyunova$^{1}$,
    Roei Herzig$^{3}$,
    Zexue He$^{2}$,
    Zihan Wang$^{4}$,
    Xinyue Yu$^{4}$,
    Yunfei Zhao$^{4}$,
    Sicong Jiang$^{4}$,
    Minghao Liu$^{4}$,
    Qunshu Lin$^{4}$,
    Peter Staar$^{3}$,\\
    Luis Lastras$^{3}$,
    Aude Oliva$^{1,2}$,
    Rogerio Feris$^{2,3}$\thanks{Correspondence: \tt rsferis@us.ibm.com}
    \\[0.5em]
    {\small $^{1}$MIT \quad $^{2}$MIT-IBM Watson AI Lab \quad $^{3}$IBM Research \quad $^{4}$Abaka AI \& 2077AI}
  }
}

\begin{document}
 \maketitle
\input{sec/0_abstract}    
\input{sec/1_intro}
\input{sec/2_related_work}

\input{sec/3_dataset}

\input{sec/4_experiments}
\input{sec/5_discussion}

\input{sec/6_conclusion}
\input{sec/7_acknowledgements}

{
    \small
    \bibliographystyle{ieeenat_fullname}
    \bibliography{main}
}

\appendix
\input{sec/X_suppl}

\end{document}

%% file: sec/0_abstract.tex
\begin{abstract}
Understanding charts requires models to jointly reason over geometric visual patterns, structured numerical data, and natural language — a capability where current vision-language models (VLMs) remain limited. We introduce ChartNet, a high-quality, million-scale multimodal dataset designed to advance chart interpretation and reasoning. ChartNet leverages a novel code-guided synthesis pipeline to generate 1.5 million diverse chart samples spanning 24 chart types and 6 plotting libraries. Each sample consists of five aligned components: plotting code, rendered chart image, data table, natural language summary, and question-answering with reasoning, providing fine-grained cross-modal alignment. To capture the full spectrum of chart comprehension, ChartNet additionally includes specialized subsets encompassing human annotated data, real-world data, safety, and grounding. Moreover, a rigorous quality-filtering pipeline ensures visual fidelity, semantic accuracy, and diversity across chart representations. 
Fine-tuning on ChartNet consistently improves results across benchmarks, demonstrating its utility as large-scale supervision for multimodal models. As the largest open-source dataset of its kind, ChartNet aims to support the development of foundation models with robust and generalizable capabilities for data visualization understanding. The dataset is publicly available at
\href{https://huggingface.co/datasets/ibm-granite/ChartNet}{https://huggingface.co/datasets/ibm-granite/ChartNet}.
\end{abstract}

%% file: sec/1_intro.tex
\section{Introduction}
\label{sec:intro}

Charts are a fundamental medium for communicating quantitative information across scientific, financial, and business domains. They translate structured data into visual form, allowing readers to efficiently reason about trends, distributions, and relationships. However, interpreting such visualizations requires integration of visual, numerical, and linguistic understanding -- a capability that current vision–language models (VLMs) only partially achieve.

Despite a growing body of work on chart understanding and reasoning, progress remains bounded by data limitations. Existing datasets are often limited in size, narrow in scope, or incomplete in their multimodal coverage. Many focus on a single task (e.g., question answering or captioning) or lack critical modalities such as plotting code, grounding annotations, or reasoning traces. Consequently, open-source models continue to lag behind proprietary systems in complex chart reasoning tasks that demand tight coupling between visual perception, structured data extraction, and natural language interpretation.

To address this gap, we introduce \textit{ChartNet}, a million-scale, high-quality multimodal dataset designed to advance robust chart understanding. ChartNet builds on a code-guided synthetic generation pipeline capable of producing chart tuples at scale that jointly capture the visual, structural, numerical, and textual aspects of chart understanding. Each instance in the dataset includes a rendered chart image, executable plotting code, underlying data table, natural-language summary, and question-answering with reasoning, ensuring complete modality alignment and interpretability. In addition, ChartNet incorporates real-world and human-annotated data, as well as specialized subsets supporting grounding, and safety analysis -- broadening the dataset’s utility for both model training and evaluation.

We perform a thorough experimental analysis, and demonstrate the value of ChartNet across models of various sizes on multiple chart understanding tasks. We also find that our best finetuned model outperforms models order-of-magnitude larger as well as GPT-4o across all tasks.

\begin{table*}[htbp]

\centering
\small
\setlength{\tabcolsep}{2.5pt}
\begin{tabular}{@{}lccccccccccccc@{}}
\toprule
\textbf{Dataset} &
\textbf{\#Charts} &
\textbf{\#Types} &
\textbf{\#Plot Libs} &
\textbf{Real} &
\textbf{Code} &
\textbf{Data} &
\textbf{Text} &
\textbf{QA} &
\textbf{Reason.} &
\textbf{Human} &
\textbf{Grounding} &
\textbf{Safety} &
\textbf{Task} \\
\midrule
FigureQA \cite{kahou2018figureqa}         & 100K  & 3     & 1            & \xmark & \xmark & \xmark & \xmark & \cmark & \xmark & \xmark & \xmark & \xmark & Binary QA \\
DVQA  \cite{kafle2018dvqa}                & 300K  & 1     & 1            & \xmark & \xmark & \xmark & \xmark & \cmark & \xmark & \xmark & \xmark & \xmark & Fixed QA \\
PlotQA  \cite{methani2019plotqa}          & 224K  & 4     & 1            & \cmark & \xmark & \xmark & \xmark & \cmark & \xmark & \xmark & \xmark & \xmark & Open QA \\
ChartQA  \cite{masry2022chartqa}          & 14K   & 3     & 1            & \cmark & \xmark & \xmark & \xmark & \cmark & \xmark & \cmark & \xmark & \xmark & Complex QA \\
Chart-to-Text  \cite{kantharaj2022charttotext} & 44K  & Multi & 1         & \cmark & \xmark & \xmark & \cmark & \xmark & \xmark & \cmark & \xmark & \xmark & Summary \\
OpenCQA  \cite{kantharaj2022opencqa}      & 7.7K  & Multi & 1            & \cmark & \xmark & \cmark & \cmark & \cmark & \xmark & \cmark & \xmark & \xmark & Long QA \\
UniChart  \cite{masry2023unichart}        & 611K  & 3     & Multi   & \cmark & \cmark & \cmark & \cmark & \cmark & \xmark & \cmark & \xmark & \xmark & Multi-task \\
ChartLlama \cite{han2023chartllama}       & 7.8K  & 10+   & 1            & \xmark & \cmark & \xmark & \cmark & \cmark & \xmark & \xmark & \xmark & \xmark & Multi-task \\
StructChart \cite{xia2024structchart}     & 9K    & 3     & 1            & \xmark & \cmark & \cmark & \xmark & \xmark & \xmark & \xmark & \xmark & \xmark & Structure \\
MMC \cite{liu2024mmc}                     & 600K  & 6     & Multi   & \cmark & \xmark & \cmark & \cmark & \cmark & \xmark & \cmark & \xmark & \xmark & Instruct \\
ChartX   \cite{xia2025chartxchartvlm}     & 6K    & 18    & 4+           & \xmark & \cmark & \cmark & \cmark & \cmark & \xmark & \cmark & \xmark & \xmark & Multi-task \\
TinyChart \cite{zhang2024tinychart}       & 680K  & Multi & Multi   & \cmark & \xmark & \xmark & \cmark & \cmark & \cmark & \cmark & \xmark & \xmark & QA+Summary \\
ChartQAPro  \cite{masry2025chartqapro}    & 1.3K  & Multi & 1            & \cmark & \xmark & \xmark & \xmark & \cmark & \cmark & \xmark & \xmark & \xmark & Diverse QA \\
Plot2Code \cite{wu2024plot2code}          & 132   & 6     & 1            & \xmark & \cmark & \xmark & \xmark & \xmark & \cmark & \xmark & \xmark & \xmark & Code Gen \\
ChartMimic \cite{yang2025chartmimic}      & 4.8K  & 22    & Multi   & \cmark & \cmark & \xmark & \xmark & \xmark & \xmark & \cmark & \xmark & \xmark & Chart-to-code \\
ChartCoder \cite{zhao2025chartcoder}      & 160K  & \textbf{27} & 1      & \xmark & \cmark & \xmark & \xmark & \xmark & \xmark & \xmark & \xmark & \xmark & Code Gen \\
CoSyn  \cite{yang2025scaling}             & 118K  & 22    & 3+           & \xmark & \cmark & \cmark & \cmark & \cmark & \cmark & \xmark & \xmark & \xmark & Multi-task \\
\midrule
\rowcolor{violet!10}
\textbf{ChartNet (ours)} 
                  & \textbf{1.5M}
                  & 24
                  & \textbf{6}
                  & \textbf{\cmark}
                  & \cmark
                  & \cmark
                  & \cmark
                  & \cmark
                  & \cmark
                  & \cmark
                  & \textbf{\cmark}
                  & \cmark
                  & \textbf{Multi-task} \\
\bottomrule

\end{tabular}

\caption{Comparison of chart understanding datasets. We report the number of unique chart images, chart types, and plotting libraries included; types of data modalities included—real-world charts/data, plotting code, tabular data, text descriptions, question-answer pairs, reasoning traces, human annotations, grounding signals, and safety data— and the scope of chart understanding tasks covered.}

\label{tab:chart_datasets}
\end{table*}

Our contributions are threefold:

\begin{enumerate}
    \item We propose a code-guided automatic chart generation pipeline that integrates structured data synthesis with automated quality filtering, ensuring visual fidelity, semantic correctness, and representational diversity at scale.
    \item We release \textit{ChartNet}, the largest-to-date synthetic chart dataset, spanning diverse chart types, plotting libraries, and topics. It contains 1.5 million high-quality multimodal tuples (image, code, CSV, text, and reasoning-based QA), as well as subsets including human annotations, grounding, safety data, and real-world charts.
    \item We demonstrate the utility of ChartNet through comprehensive experiments, showing that finetuning on this dataset consistently improves chart reconstruction, data extraction, and chart summarization performance across vision–language models.
\end{enumerate}

ChartNet establishes a new standard for multimodal chart understanding by unifying scale, diversity, and representational completeness, enabling the next generation of models to reason over data visualizations with greater accuracy and generalization.

%% file: sec/2_related_work.tex
\section{Related Work}
\label{sec:related_work}

\begin{figure*}[htbp]
    \centering
    \includegraphics[width=1.0\linewidth]{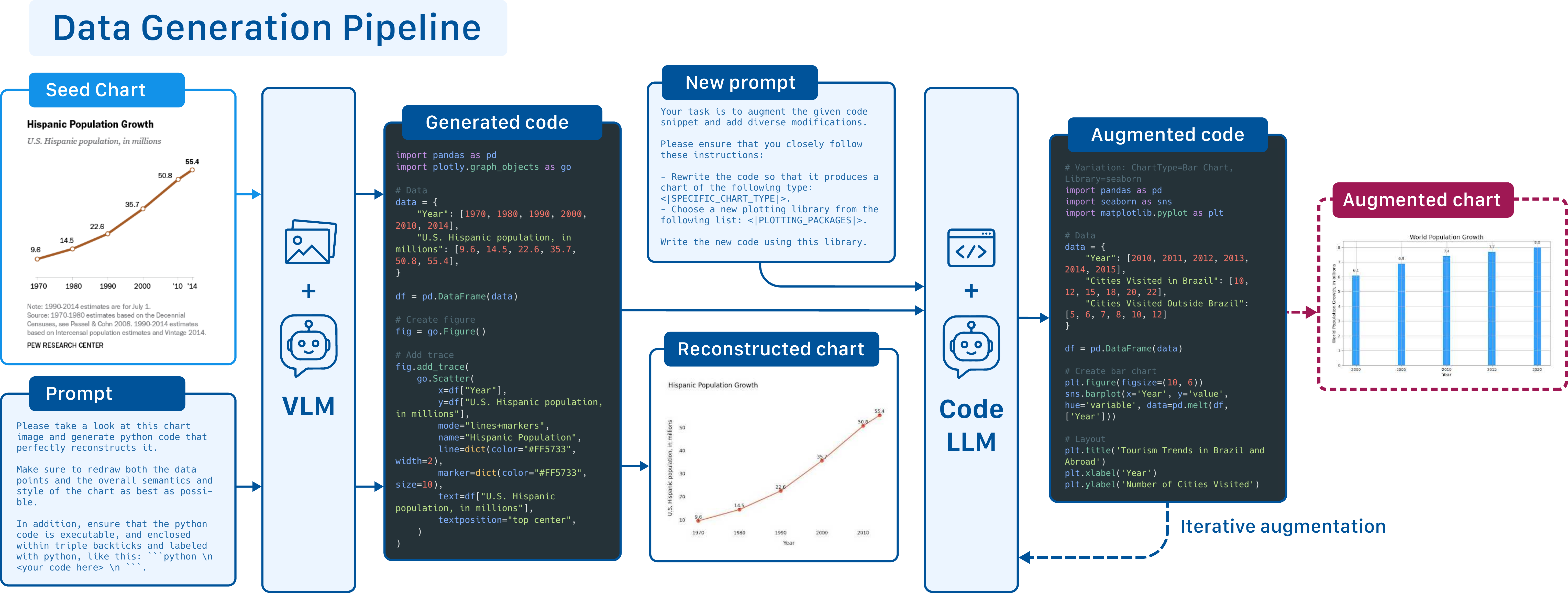}
    \caption{Code-guided chart augmentation: First, a seed chart image is passed to a vision-language model for chart reconstruction -- translating the image into executable plotting code. Then, the generated code is passed to a large language model, and iteratively augmented to collect diverse outputs.}
    \label{fig:pipeline}
\end{figure*}

\subsection{Large Multimodal Models.} Open-source multimodal models \cite{qwen3technicalreport,team2025kimi,liu2023improvedllava,an2025llava,hong2025glm} have made notable progress on document and chart comprehension benchmarks, yet their performance generally falls short of leading proprietary models.  Recent efforts to close this gap include architectural improvements, such as enhanced high-resolution image processing \cite{guo2024llava,hua2025finecaption,zhu2025internvl3} and explicit numerical reasoning \cite{zhang2024tinychart,sun2025latent}. Nevertheless, the scarcity of high-quality chart comprehension training data remains a critical bottleneck. This challenge is compounded by the lack of transparency surrounding data curation practices in even the best-performing open models  \cite{qwen3technicalreport,meta2025llama}, creating significant barriers to reproducibility. Our ChartNet dataset, on the other hand, provides large scale, high quality data for advancing the chart understanding capabilities of multimodal models, while being made freely available to the research community.

\subsection{Chart Understanding Datasets}

Numerous datasets have been proposed for chart question-answering \cite{kahou2018figureqa,kafle2018dvqa,methani2019plotqa,masry2022chartqa,kantharaj2022opencqa,wang2024scicqa,huang2025evochartbenchmarkselftrainingapproach}, captioning and summary generation \cite{kantharaj2022charttotext,tang2023vistext}, chart-to-code translation \cite{wu2024plot2code,yang2025chartmimic,zhao2025chartcoder,liu2025chartreasoner}, and multimodal chart reasoning \cite{han2023chartllama,xu2024chartbench,wang2024charxiv,xia2025chartxchartvlm,liu2024synchart,chen2025chartgalaxy,yang2025scaling,meng2024chartassisstant}. However, these datasets fail to capture the full diversity of real-world charts. For example, ChartQA \cite{masry2022chartqa} -- a widely used benchmark for multimodal models -- encompasses only a few chart types (bar, line, and pie charts) obtained from limited online sources. Moreover, it is biased towards questions requiring basic data extraction, resulting in performance saturation for modern vision-language models. While recent datasets have addressed some of these limitations by incorporating more realistic charts \cite{li-etal-2024-multimodal-arxiv} and more complex questions \cite{masry2025chartqapro}, they still lack the diversity, scale, and quality required to train frontier large multimodal models. In contrast, ChartNet is a million-scale dataset featuring 24 different chart types and various plotting libraries, with rigorous data filtering, high-quality human annotations, and associated tasks including chart-to-code, chart data extraction, chart captioning, reasoning, grounding, and safety. Table \ref{tab:chart_datasets} compares ChartNet with other datasets.







\subsection{Synthetic Data Generation for Vision-Language Models}

Recently, synthetic data generation has gained significant attention from both industry and academia as an effective means to improve the capabilities of VLMs \cite{zhu2023multimodal,hua2025v2xum,guo2025mammoth}. It has proven especially valuable for tasks such as visual question answering \cite{antol2015vqa,kembhavi2016diagram,marino2019ok,hua2025mmig} and compositional reasoning \cite{hua2024mmcomposition,johnson2017clevr,hudson2019gqa,hua2024finematch,tang2025vidcomposition}. In contrast, our approach performs data generation and augmentation in the code space as opposed to the image space. Granite Vision \cite{team2025granite}, DAVE \cite{huang2025dave}, SmolDocling \cite{docling}, Molmo \cite{deitke2024molmo}, and CoSyn \cite{yang2025scaling} also rely on synthetic data generation for charts and documents tasks. Different from our work, they exhibit limited diversity in chart types and modalities compared to ChartNet.

%% file: sec/3_dataset.tex
\section{ChartNet Data Generation Pipeline}
\label{subsec:pipeline}
A key methodological insight underlying our data generation is that charts are generated
programmatically: executable plotting code serves a structured intermediate representation for
data visualizations \cite{kondic2025chartgen}.
We introduce an automated pipeline for code-guided synthetic chart generation at scale (see Figure \ref{fig:pipeline}). 
Starting with a limited dataset of chart images ("seeds"), a VLM outputs code that approximately reconstructs them. We then leverage the code representation to (1) iteratively generate augmentations, producing visually and semantically diverse charts, and (2) generate additional semantic attributes, including tabular data, natural language descriptions, and question-answering traces with chain-of-thought reasoning.

\begin{figure*}[htbp]
  \centering
  \includegraphics[width=1.0\linewidth]{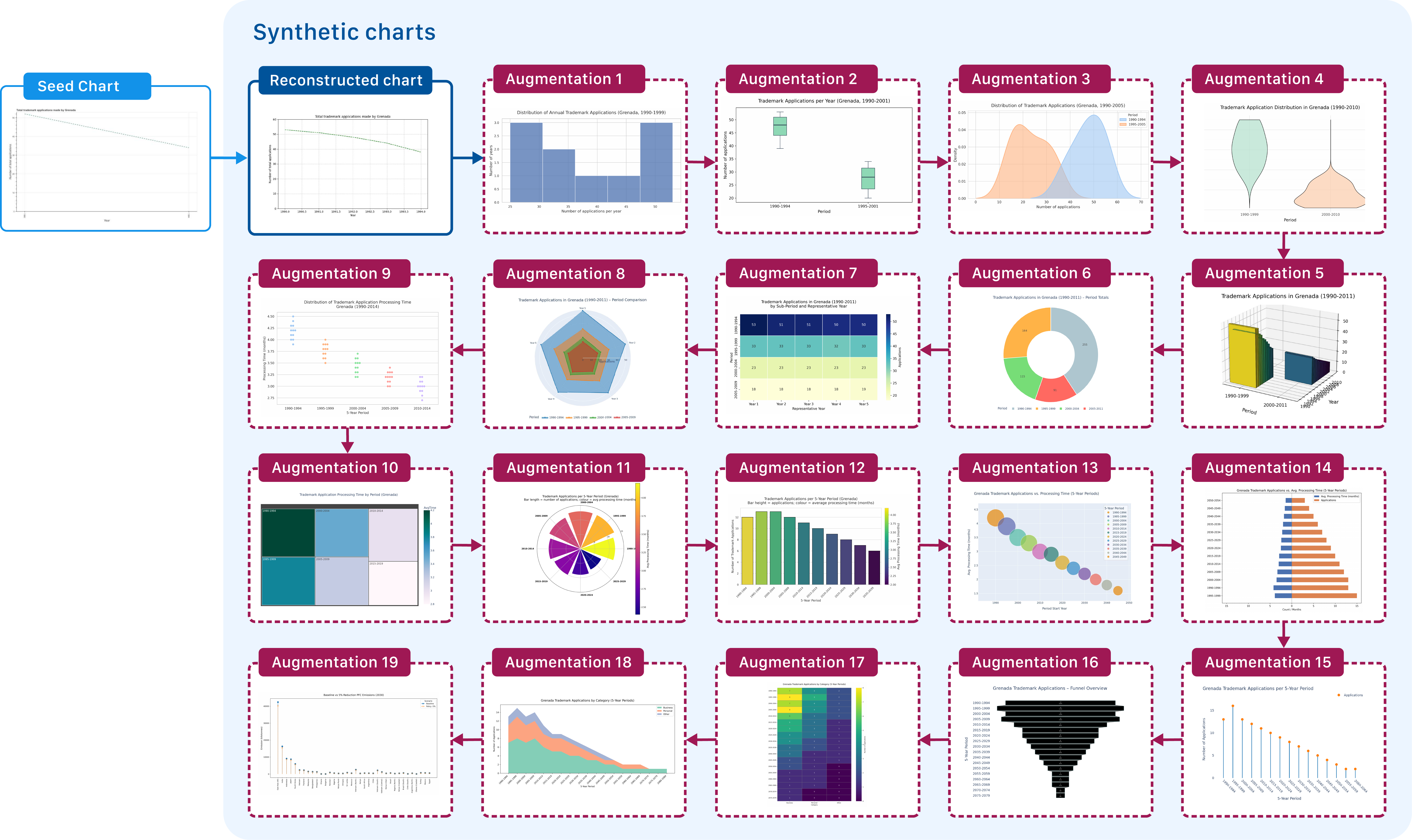} 
  \caption{An illustration of synthetic chart images generated from a single seed chart using the ChartNet pipeline. A seed chart is first translated into approximate plotting code, which is executed to render a reconstructed chart. The code is then iteratively augmented to produce diverse variations in chart types, styles, and representations, as shown in the subsequent \textit{augmentations}.}
  \label{fig:augmentation process}
\end{figure*}

\subsection{Code-Guided Data Generation At Scale}
Specifically, our data generation pipeline consists of the following stages:

\begin{enumerate}

    \item \textbf{Chart-to-Code Reconstruction}: 
    We prompt a VLM to produce Python plotting code that approximately reconstructs a given set of chart images. Here, we select a seed set of $150,000$ unique chart images from TinyChart \cite{zhang2024tinychart}, though the pipeline is agnostic to this choice. 
    \item \textbf{Code-Guided Chart Augmentation}: Using the produced plotting code as input, we prompt an LLM to iteratively rewrite it. The underlying data values and labels are transformed to better match the requested chart type, while maintaining relevance to the previous iteration. Figure \ref{fig:augmentation process} illustrates the iterative code augmentation and chart rendering process. This stage is the primary contributor to dataset scaling, taking each seed image and producing up to an arbitrary number of variations. 
    \item \textbf{Chart Rendering}: We execute all the generated plotting code to produce chart images. The scripts that were successful upon execution are paired with the image that they produced.
    \item \textbf{Quality Filtering}: Using a VLM, we evaluate each chart image across multiple identified categories of potential rendering defects (e.g., overlapping text, cropped labels, obscured features). 
    Images classified with visual issues (and their plotting code) are removed.
    \item \textbf{Code-Guided Attribute Generation}: Finally, we use a VLM to generate supplementary semantic attributes to the chart image-code pairs.
    Grounding the visual information with code as additional context, we extract the data values and labels from charts and produce tabular data representations. 
    Furthermore, combining the visual context with code and tabular data, we produce grounded chart descriptions.

\end{enumerate}

For prompt templates used, see Section \ref{supp_data_gen_prompts}.

\subsection{QA Pairs with CoT Reasoning} 
In addition to chart image, code, tabular data, and natural language descriptions, we also generate question-answer (QA) pairs with long Chain-of-Thought (CoT) reasoning as part of the ChartNet dataset. This data generation process is built on the Vision-R1 framework \cite{huang2025vision}. Using pixtral-large-instruct-2411, we generate a complex multi-stage reasoning question for each image in the ChartNet dataset. Next, following the procedure proposed in LLaVA-CoT \cite{xu2025llava}, we construct a four-step “Pseudo-CoT” sequence (Summary, Caption, Reasoning, and Conclusion) using separate model calls. We then perform modality bridging, where the model describes the complete visual content in relation to the Pseudo-CoT, enabling a language-only model to reason effectively without direct visual input. Finally, gpt-oss-120b \cite{agarwal2025gpt} produces detailed textual reasoning traces and final predictions enclosed within \texttt{<think>} and \texttt{<answer>} tags.
This multi-stage pipeline produces rich, verifiable reasoning traces while preserving strong alignment between visual and textual representations. 
See Section \ref{sec:cot-prompts} for more information and illustrative examples.

\begin{figure*}[ht!]
  \centering
  \includegraphics[width=0.92\linewidth]{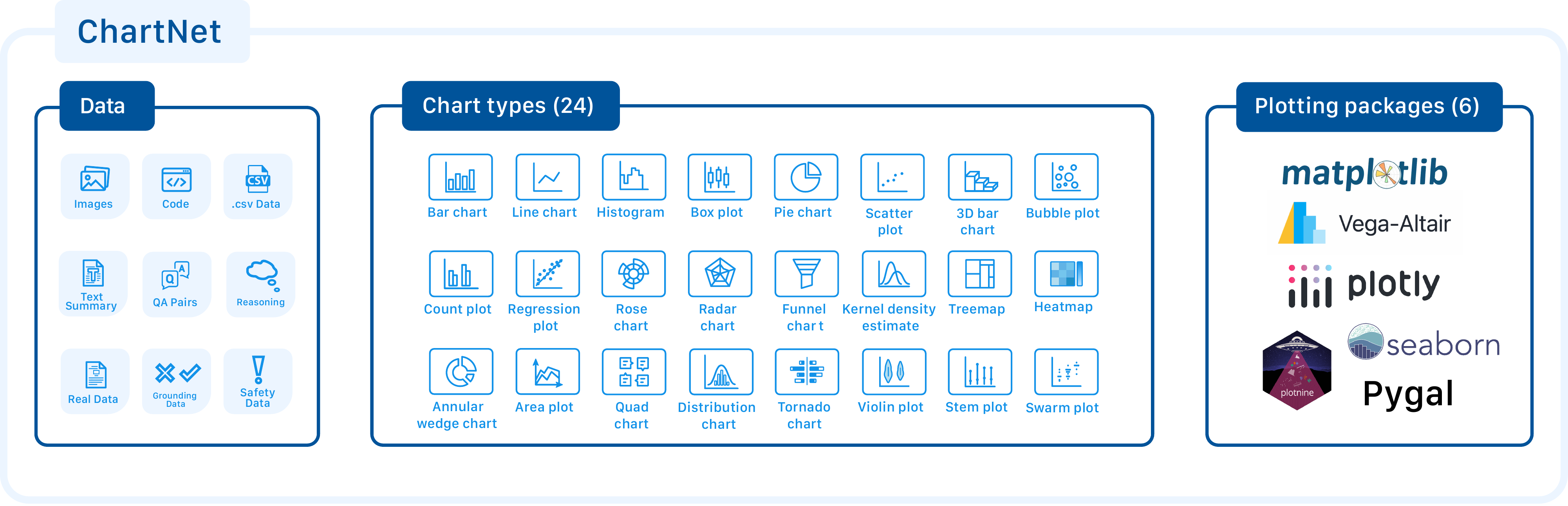}
  \caption{Data attributes, chart types, and plotting packages included in ChartNet.}
  \label{fig:chart net}
\end{figure*}

\subsection{Models and Compute Infrastructure}~\label{modelscompute}
Our model choice was based on a combination of demonstrated performance and adhering to open-source values.
We use pixtral-large-instruct-2411 in the Chart-to-Code Reconstruction, Quality Filtering, and Code-Guided Attribute Generation stages, and gpt-oss-120b in the Code-Guided Chart Augmentation stage. For scale, we deployed multiple replicas of both models on over a hundred A100 and H100 GPUs. The work was distributed across the GPUs to maintain high throughput, generating over 1 million annotated data points roughly every 168 hours. 


\subsection{Quality Filtering Evaluation}
In the Quality Filtering stage, we track three observable metrics across three stages, and observe the following: 
\begin{itemize}
    \item \textbf{Probability of Failure} (Chart Augmentation): The model fails to rewrite the code snippet with requested changes and proper formatting in $<$0.01\% of requests.
    \item \textbf{Execution Rate} (Chart Rendering): On average, 77\% of the generated code snippets execute successfully.
    \item \textbf{Visual Error Rate} (Quality Filtering): On average,  36.5\% of rendered images were classified to contain some visual error.
\end{itemize}

To quantify how well pixtral-large-instruct-2411 aligns with human performance in detecting visual defects, 3157 randomly sampled charts were manually annotated and compared to the corresponding model prediction. Before Quality Filtering, 14.9\% of generated samples were found to contain issues that affect chart readability. After Quality Filtering, only 5.9\% of the charts contained these issues. 

%


\section{The ChartNet Dataset}
\label{sec:dataset}
\subsection{Core Dataset}
\label{subsec:core data}
The core ChartNet dataset consists of 1.5M multimodal aligned synthetic tuples: chart image, plotting code, tabular data, natural language description, and QA pairs with CoT reasoning. 
For a complete overview of the data attributes, chart types, and plotting packages included, see Fig.~\ref{fig:chart net}.




To capture the full spectrum of chart understanding, ChartNet additionally includes specialized subsets: human-annotated data, real-world charts, grounding, and safety.

\subsection{Human-Annotated Synthetic Chart Data}

In addition to the core dataset, we curate a high-quality subset of $96,643$ aligned synthetic chart images, descriptions, and tabular data that have gone through rigorous human verification and annotation. See Section~\ref{humanannotation}  for more information about the annotation process.

\subsection{High-Quality Real-World Charts}

To complement our synthetic chart corpus, we also curate and annotate 30K real-world charts sourced from reputable international media and data-visualization outlets such as the \textit{World Bank} \cite{worldbank2025data}, \textit{Bain Insights} \cite{baininsights2025}, \textit{Pew Research Center} \cite{pew2025}, \textit{Our World in Data} \cite{owid2025}, and other globally recognized publishers. This collection captures a broad spectrum of contemporary topics, including economics, technology, geopolitics, environmental science, and societal trends, also ensuring high diversity and strong real-world relevance. We explicitly discard a broad set of low-information or low-quality visuals that do not meet our interpretability standard. To ensure full compliance with copyright and data-use regulations, all real-world charts 
were collected exclusively from legally safe, openly licensed, or public-domain sources, and their use falls strictly under non-commercial academic research exceptions.


Each selected chart is paired with metadata, including its caption, sub-caption, key data highlights, and a concise analytical summary, to support joint learning of visual reasoning, textual grounding, and high-level insight extraction.
This subset is specifically designed to strengthen multimodal model performance on challenging chart understanding tasks, including:
\begin{itemize}
    \item \textbf{Quantitative and comparative reasoning}: extracting values, trends, anomalies, and multi-series comparisons directly from visual structures;
    \item \textbf{Chart--text semantic alignment}: linking visual elements with captions, labels, and narrative descriptions;
    \item \textbf{Context-aware summarization}: generating coherent explanations that integrate both visual evidence and accompanying textual information;
    \item \textbf{Cross-lingual interpretation}: supporting multilingual understanding of globally sourced visualizations.
\end{itemize}

For additional information and illustrative examples, see Section
\ref{appendix:real-world}.

\subsection{Grounding QA Pairs}

Modern VLMs still struggle to identify the chart areas and syntactic elements relevant to a given question. To further advance such capabilities, we create grounding QA pairs. First, we extract geometry-aware annotations from elements of the plotting code (axes, ticks, gridlines, legends, patches) to produce dense grounding annotations of the corresponding charts. Bounding boxes are further filtered using an entropy-based approach (see Section ~\ref{bbox}).
Using the resulting grounded annotations, for each chart, we create a set of template-based QAs that 
capture the duality between the expected spatial arrangement of visual elements and the observed content depicted in the plots. The expected locations are encoded as serialized bounding-boxes representations 
within the corresponding answer strings.

Templates address unique and recurring visual elements, incorporating referring expressions based on indices, textual labels present in the plot, and visual attributes (e.g., element color).
The generator supports both short- and long-form answers, and can optionally include grounding information for each.
The final dataset is obtained by uniform sampling across all template types and output modalities, generating one QA pair per image. In addition to this, we include a set of reasoning-based grounding QA pairs by leveraging gpt-oss-120b. Section ~\ref{subsec:grounding} provides more information and points to examples of the generated QA pairs.




\subsection{Safety}
To address safety concerns, we extend our pipeline to generate chart-related safety alignment data that mitigates harmful model outputs and jailbreak vulnerabilities. We first select charts with sensitive content across topics including health, finance, and social issues. We then synthetically generate adversarial questions spanning categories such as discrimination, hate, violence, political bias, and substance abuse (e.g., "Does this bar chart prove that Race X causes higher crime rates?"). Each question is paired with both safe and unsafe responses, creating preference pairs suitable for direct preference optimization. We release 7,000 training samples and 600 test samples as part of ChartNet. For prompt templates and more information, see Section \ref{appendix: safety}.



%% file: sec/4_experiments.tex
\section{Experiments}
\label{sec:experiments}

\subsection{Model Training}
\label{subsec:training}

We train VLMs of various sizes on the ChartNet dataset to validate its effectiveness in enhancing models’ chart understanding capabilities. The supervised finetuning (SFT) data comprises the four tasks of the core ChartNet dataset: \textbf{Chart-to-Code}, \textbf{Chart-to-Table}, \textbf{Chart-to-Text}, and \textbf{Chart QA with CoT Reasoning}. Specifically, we experiment with different model scales: Ultra-Compact ($\leq$1B) — Granite-Docling-258M \cite{livathinos2025docling} and SmolVLM-256M-\cite{marafioti2025smolvlm}; Small ($\leq$4B) — Granite-vision-3.3-2b \cite{team2025granite} and Qwen2.5-VL-3B-Instruct \cite{bai2025qwen2}; and Medium ($\leq$7B) — LLaVA-v1.6-mistral-7b \cite{ liu2023improvedllava}. We follow the default hyperparameter settings provided by the TRL\cite{vonwerra2022trl} codebase.


\begin{table*}[htbp]
\centering
\resizebox{0.98\textwidth}{!}{
\begin{tabular}{l | c c c c | c | c | c}
\toprule
\textbf{Model (params)} 
& \multicolumn{4}{c|}{\textbf{Chart Reconstruction }} 
& \multirow{2}{*}{\shortstack{\textbf{Chart Data}\\\textbf{Extraction}}}
& \multirow{2}{*}{\shortstack{\textbf{Chart}\\\textbf{Summarization}}}
& \multirow{2}{*}{\shortstack{\textbf{Chart QA w/}\\\textbf{CoT Reasoning}}} \\
\cmidrule(lr){2-5}
& Exec. & Code-D & Code-S & Img. &  &  &  \\
\midrule

\texttt{SmolVLM-Instruct (256 M)} 
& N/A & N/A & N/A & N/A & 22.0 & 26.6 & 55.0 \\
\rowcolor{violet!10}
\texttt{+ ChartNet } 
& 14.9 & 56.8 & 74.9 & 77.5 & 36.4 &  60.0 & 60.8 \\
\textcolor{blue}{(+)}
& \textcolor{blue}{(+14.9)} & \textcolor{blue}{(+56.8)} & \textcolor{blue}{(+74.9)} & \textcolor{blue}{(+77.5)} & \textcolor{blue}{(+14.4)} & \textcolor{blue}{(+33.4)} & \textcolor{blue}{(+5.8)} \\
\midrule

\texttt{granite-docling-258M (258 M)} 
& N/A & N/A & N/A & N/A & 17.7 & 24.1 & 54.0 \\
\rowcolor{violet!10}
\texttt{+ ChartNet} 
& 41.8 & 49.7 & 63.1 & 70.4 & 32.0 & 55.5 & 61.0 \\
\textcolor{blue}{(+)} 
& \textcolor{blue}{(+41.8)} & \textcolor{blue}{(+49.7)} & \textcolor{blue}{(+63.1)} & \textcolor{blue}{(+70.4)} 
& \textcolor{blue}{(+14.3)} & \textcolor{blue}{(+31.4)} & \textcolor{blue}{(+7.0)} \\
\midrule

\texttt{granite-vision-3.3-2B (2 B)} 
& 63.4 & 60.7 & 67.0 & 77.2 & 53.8 & 64.0 & 59.9 \\
\rowcolor{violet!10}
\texttt{+ ChartNet } 
& \textbf{90.4} & \textbf{72.8} & \textbf{90.0} & \textbf{92.8} & \textbf{70.3} & \textbf{83.9} & 65.0 \\
\textcolor{blue}{(+)}
& \textcolor{blue}{(+27.0)} & \textcolor{blue}{(+12.1)} & \textcolor{blue}{(+23.0)} & \textcolor{blue}{(+15.6)} 
& \textcolor{blue}{(+16.5)} &  \textcolor{blue}{(+19.9)} & \textcolor{blue}{(+5.1)} \\
\midrule

\texttt{Qwen2.5-VL-3B-Instruct (3 B)} 
& 65.1 & 52.6 & 68.0 & 76.7 & 51.8 & 70.6 & 58.4 \\
\rowcolor{violet!10}
\texttt{+ ChartNet} 
& 74.1 & 59.1 & 76.1 & 82.8 & 62.4 & 80.1 & 69.9 \\
\textcolor{blue}{(+)} 
& \textcolor{blue}{(+9.0)} &  \textcolor{blue}{(+7.5)} & \textcolor{blue}{(+8.1)} & \textcolor{blue}{(+6.1)} & \textcolor{blue}{(+10.6)} & \textcolor{blue}{(+9.5)} & \textcolor{blue}{(+11.5)} \\
\midrule

\texttt{llava-v1.6-mistral-7b-hf (7 B)} 
& 45.3 & 27.0 & 52.9 & 59.6 & 17.0 & 51.2 & 55.1 \\
\rowcolor{violet!10}
\texttt{+ ChartNet} 
& 83.9 & 69.4 & 88.6 & 91.5 & 58.8 & 80.3 & \textbf{70.3} \\
\textcolor{blue}{(+)}
& \textcolor{blue}{(+38.6)} & \textcolor{blue}{(+42.4)} & \textcolor{blue}{(+35.7)} & \textcolor{blue}{(+31.9)} 
& \textcolor{blue}{(+41.8)} & \textcolor{blue}{(+29.1)} & \textcolor{blue}{(+15.2)} \\

\bottomrule
\end{tabular}
}

\caption{Paired comparison of base models vs finetuned models on the ChartNet evaluation set, with performance gains in blue \protect\footnotemark. Each model variant was trained solely on the subset of the ChartNet dataset corresponding to the specific task it was evaluated on (for example, models evaluated on Chart Reconstruction were trained only on the Chart-to-Code subset of ChartNet).}

\label{tab:data_utility}
\end{table*}

\afterpage{\footnotetext{Latest updates and enhancements to the dataset are available at \url{https://huggingface.co/datasets/ibm-granite/ChartNet}. See also results using \texttt{granite-4.0-3b-vision} finetuned on ChartNet at \url{https://huggingface.co/ibm-granite/granite-4.0-3b-vision}.}}

\begin{table*}[htbp]
\centering
\resizebox{\textwidth}{!}{
\begin{tabular}{l | c c c c | c | c | c}
\toprule
\textbf{Model (params)} 
& \multicolumn{4}{c|}{\textbf{Chart Reconstruction}} 
& \multirow{2}{*}{\shortstack{\textbf{Chart Data}\\\textbf{Extraction}}}
& \multirow{2}{*}{\shortstack{\textbf{Chart}\\\textbf{Summarization }}}
& \multirow{2}{*}{\shortstack{\textbf{Chart QA w/}\\\textbf{CoT Reasoning}}} \\
\cmidrule(lr){2-5}
& Exec. & Code-D  & Code-S & Img. &  &  &  \\
\midrule
\rowcolor[HTML]{e9edf6}
\multicolumn{8}{c}{\textbf{Open-Source Models}} \\
\midrule
\texttt{Qwen3-VL-3B-Instruct (3 B)}               & 76.9 & \textbf{63.3} & 74.3 & 86.8 & \textbf{58.1} & 79.2 & 64.3\\
\texttt{InternVL3\_5-8B (8 B)}                    & 69.1 & 60.2 & 69.9 & 78.0 & 56.1 & 71.3 & 61.6 \\
\texttt{Pixtral-12B-2409 (12 B)}                  & 74.9 & 52.9 & 72.9 & 81.4 & 49.1 & 77.5 & 60.0 \\
\texttt{Mistral-Small-3.1-24B-Instruct-2503 (24 B)} & 88.1 & 54.5 & 74.8 & 86.3 & 53.2 & \textbf{79.8} & 60.0 \\
\texttt{Qwen2-VL-72B-Instruct (72 B)}             & 83.1 & 50.7 & 67.3 & 77.6 & 50.3 & 75.9 &  60.3 \\
\midrule
\rowcolor[HTML]{e9edf6}
\multicolumn{8}{c}{\textbf{Chart Models}} \\
\midrule
\texttt{ahmed-masry/chartgemma (3B)}              &  & N/A & &  & 37.1 & 46.1 & \textbf{69.5}  \\
\midrule
\rowcolor[HTML]{e9edf6}
\multicolumn{8}{c}{\textbf{Proprietary Models}} \\
\midrule
\texttt{GPT-4o}                                   & \textbf{95.9} & 48.8 & \textbf{77.2} & \textbf{88.2} & 46.7 & 77.1 & 61.1 \\
\bottomrule
\end{tabular}
}
\caption{Performance of off-the-shelf models on the ChartNet evaluation set.}
\label{tab:other_models}

\end{table*}

\subsection{ChartNet Evaluation Set}

To rigorously evaluate the tasks in the core ChartNet dataset, we curate a held-out evaluation suite randomly drawn from ChartNet’s synthetic corpus. The set comprises 2000 chart tuples, each including a chart image, its corresponding plotting code, underlying data table, a natural language summary, and QA pairs with CoT reasoning.
We evaluate model performance across four tasks:

\vspace{-3mm}
\paragraph{Chart Reconstruction (Chart-to-Code).}
Given a chart image $I$, the model is required to generate an executable plotting script $C'$ that reproduces as closely as possible the source code $C$ used to render the input chart $I$. We evaluate (a) \textit{execution rate} (Exec.) — the fraction of generated scripts $C'$ that execute without error, (b) \textit{data fidelity} (Code-D) — the correspondence between plotted numeric values and the data defined in ground-truth code, (c) \textit{code similarity} (Code-S) — the structural and syntactic overlap between generated, $C'$, and source code, $C$, and (d) \textit{rendered image similarity} (Img.) — the visual alignment between the rendered prediction and the input chart $I$.

\vspace{-3mm}
\paragraph{Chart Data Extraction (Chart-to-Table).}
This task evaluates the ability of a model to infer the plotted data directly from the chart image. Given an input image $I$, a model is asked to produce a CSV table that matches as closely as possible the data points visualized in $I$. Using $I$ as context, we compare the generated data table to the ground-truth CSV, and report a similarity score disregarding minor formatting differences.

\vspace{-3mm}
\paragraph{Chart Summarization (Chart-to-Text).}
Given a chart image $I$, the model is tasked with generating a comprehensive textual summary capturing the key takeaways, data trends, comparisons, and visual elements and style of the chart. Using $I$ as context, we compare the generated summary to the reference summary generated and verified by the ChartNet data generation pipeline as described in Section \ref{subsec:pipeline}. We report a holistic score encompassing the coverage of key elements, faithfulness to the visual, semantic and numeric correctness, and clarity.

\vspace{-3mm}
\paragraph{Chart QA with CoT Reasoning}
For each chart image $I$, we pair the generated complex reasoning question with $I$, and prompt the model to output \texttt{<think>} and \texttt{<answer>} sections. The final answer is extracted from \texttt{<answer>} and compared to the gold reference using RapidFuzz for fuzzy string matching. We report average fuzzy accuracy.
\\

\noindent We evaluate a range of off-the-shelf open-source VLMs ($<1$B – $72$B parameters), a specialized chart model (ChartGemma \cite{masry2024chartgemma}), and GPT-4o, and compare these against models finetuned on ChartNet (as outlined in Section \ref{subsec:training}). All metrics are automatically computed using GPT-4o as a judge, except for the Chart QA with CoT Reasoning task. The prompt templates used are listed in Section \ref{prompts:evals}.


\subsection{Public Benchmarks}
We additionally evaluate ChartNet on established public benchmarks including chart summarization (ChartCap~\cite{lim2025chartcap}) and chart-to-code generation (ChartMimic-v2~\cite{yang2025chartmimic}). We follow the original evaluation protocols and report standard metrics, comparing both off-the-shelf models and their ChartNet-finetuned variants to prior open-source baselines.

%% file: sec/5_discussion.tex
\section{Results \& Discussion}
\label{sec:discussion}

\begin{table*}[ht!]
\centering
\setlength{\tabcolsep}{1pt}
\scalebox{0.87}{%
\begin{tabular}{@{}lccccc@{}}
Model & \multicolumn{3}{c}{ChartCap (Summarization)} & \multicolumn{2}{c}{ChartMimic-v2 (Code Generation)} \\
\cmidrule(lr){2-4}\cmidrule(lr){5-6}
 & BLEU\_4& METEOR& ROUGE\_L& v2-direct & v2-customized  \\
\midrule\specialrule{0.7pt}{0pt}{0pt}
SmolVLM-256M-Instruct & 0.60 & 14.30 & 13.50 & 0 & 0 \\
\rowcolor{violet!10}
SmolVLM-256M-Instruct-ChartNet & \textbf{3.00} & \textbf{22.00} & \textbf{17.80} & \textbf{12.67} & \textbf{11.67}  \\
\addlinespace[2pt]\specialrule{1.2pt}{0pt}{0pt}
granite-vision-3.3-2b & 1.60 & 6.40 & 9.60 & 30.84 & 30.45  \\
\rowcolor{violet!10}
granite-vision-3.3-2b-ChartNet & \textbf{12.40} & \textbf{30.10} & \textbf{24.90} & \textbf{58.42} & \textbf{51.21}  \\
\addlinespace[2pt]\specialrule{1.2pt}{0pt}{0pt}
llava-next-mistral-7b & 6.40 & 22.60 & 20.90 & 28.20 & 30.76  \\
\rowcolor{violet!10}
llava-next-mistral-7b-ChartNet & \textbf{7.10} & \textbf{27.00} & \textbf{22.10} & \textbf{54.78} & \textbf{38.27} \\
\end{tabular}%
}
\caption{Generalizability of gains from ChartNet synthetic data on two real-world public benchmarks.}
\label{tab:public_results}
\end{table*}

As shown in Table \ref{tab:data_utility}, finetuning on ChartNet produces substantial and consistent gains across all chart understanding tasks. The uniformity and magnitude of these improvements -- regardless of model scale -- indicate that existing VLMs lack exposure to high quality multimodal chart supervision, and ChartNet fills this gap effectively.

\vspace{-4mm}
\paragraph{Chart Reconstruction} Models trained on the Chart-to-Code subset show large improvements in code execution rates, data fidelity, and structural/code and image similarity. Ultra compact models that originally could not reconstruct charts at all (SmolVLM-256M, Granite-Docling-258M) gain fully functional capability, while small models such as Granite-Vision-2B achieve near-perfect reconstruction, reaching 90\%+ on most metrics. The LLaVA-7B model experiences gains of up to +42.4 points, substantially improving the code data fidelity performance. The scale-invariant trend suggests that ChartNet’s multimodal alignment between images and code provides a type of structural supervision unavailable in prior datasets.

\vspace{-3mm}
\paragraph{Chart Data Extraction} ChartNet dramatically boosts all models’ ability to recover numeric tables directly from chart images, with the best performing Granite-Vision-2B scoring 70.3\%. The finetuned LLaVA-7B model improves performance by +41.8, exceeding every open-source baseline (including those in Table \ref{tab:other_models}) and surpassing GPT-4o which shows particularly limited performance in this task, at only 46.7\% accuracy. This reflects the value of ChartNet’s tight coupling between the code-generated charts and CSVs, which gives models explicit exposure to both visual geometry and underlying numeric structure.

\vspace{-3mm}
\paragraph{Chart Summarization} Summarization quality improves across all model families, with gains ranging from +9.5 (Qwen2.5-VL-3B) to +31.4 (Granite-Docling-2B). The finetuned Granite-Vision-2B reaches 83.9\%, surpassing GPT-4o and all open-source baselines in Table \ref{tab:other_models} including those that are an order of magnitude larger. This suggests that ChartNet’s synthetic summaries, constructed jointly from code and rendered visuals, provide precisely the kind of structured, semantically complete supervision needed for descriptive chart understanding.

\vspace{-3mm}
\paragraph{QA with CoT Reasoning} Every model exhibits steady accuracy improvements on the complex multi-stage reasoning task. LLaVA-7B achieves the largest improvement (+15.17), reaching 70.3\%, outperforming a specialized chart reasoning model (ChartGemma) and all other baselines of comparable or order-of-magnitude larger sizes (including GPT-4o) in Table \ref{tab:other_models}.

\vspace{-3mm}
\paragraph{Comparison with Off-the-Shelf Models} 
Table \ref{tab:other_models} highlights that ChartNet-tuned models outperform far larger off-the-shelf models in nearly every metric. A 2B or 7B model finetuned on ChartNet consistently exceeds the performance of 20B–72B models trained on conventional multimodal corpora. In Chart Reconstruction and Chart Data Extraction, the gap is especially pronounced: ChartNet-tuned models far surpass GPT-4o overall. These results point toward an emerging principle: for domains like chart interpretation, where visual, numerical, and linguistic information are tightly coupled, scaling model size is far less effective than providing high-quality, code-aligned multimodal supervision.
\\
Collectively, these findings show the utility of ChartNet in boosting the capabilities of VLMs, enabling robust, interpretable, and numerically grounded chart reasoning that is otherwise unreachable with vision–language training.

\vspace{-3mm}
\paragraph{Generalization to Public Benchmarks} 
As shown in Table~\ref{tab:public_results}, finetuning on the core ChartNet dataset (Section \ref{subsec:core data}) yields substantial absolute gains across all models. Notably, Granite-Vision-2B improves from 1.6 to 12.4 BLEU on ChartCap, and from 30.8 to 58.4 on ChartMimic-v2, and even ultra-compact models (SmolVLM-256M) gain non-trivial capability. These improvements are consistent across both chart summarization and chart-to-code translation tasks, indicating that the aligned multimodal supervision of ChartNet transfers effectively to real-world benchmarks beyond the synthetic training distribution.


%% file: sec/6_conclusion.tex
\section{Conclusion}
\label{sec:conclusion}
We present ChartNet, addressing a central bottleneck in chart understanding: the lack of large-scale, high-fidelity supervision that aligns images, plotting code, numeric data, textual descriptions, and reasoning traces. By generating over one million aligned multimodal tuples, ChartNet equips VLMs with programmatically grounded knowledge that transfers across chart-to-code reconstruction, data extraction, summarization, and multi-step reasoning. Experiments show consistent gains across model sizes and architectures, often surpassing much larger open-source systems and even proprietary frontier models such as GPT-4o. These gains are not limited to any single task; they indicate a broader improvement in how models internalize chart semantics when trained with code-grounded supervision. ChartNet offers a scalable, open foundation for research in numerical reasoning, visualization understanding, document intelligence, and code-aligned multimodal modeling—moving VLMs from describing charts toward understanding the structured information they encode.

%% file: sec/7_acknowledgements.tex
\section*{Acknowledgments}

This work was supported in part by funding from the MIT-IBM Watson AI Lab.

%% file: sec/X_suppl.tex
\clearpage
\pagenumbering{Roman}  
\setcounter{page}{1}
\maketitlesupplementary

\section{Elaborating on Aspects of ChartNet}

\subsection{Data Distribution of the Core Dataset}

ChartNet contains a variety of charts across multiple chart types and plotting packages. During the \textit{Code-Guided Chart Augmentation} stage, we choose one of 24 different chart types uniformly randomly and ask an LLM to reformat the code in that style. While in most cases the model was able to produce a code snippet with the chosen chart type, charts of higher complexity were less likely to successfully execute due to a higher prevalence of code issues. Additionally, certain chart types were more likely to contain rendering errors (e.g., overlapping labels, obscured data) that would be flagged during the \textit{Quality Filtering} stage. As such, the distribution of chart types is not uniform. Similarly, the distribution of plotting packages is also not uniform, where code snippets generated with certain packages would execute less often, or were chosen by the model less.

\begin{figure*}[tbp]
    \centering
    \includegraphics[width=0.8\linewidth]{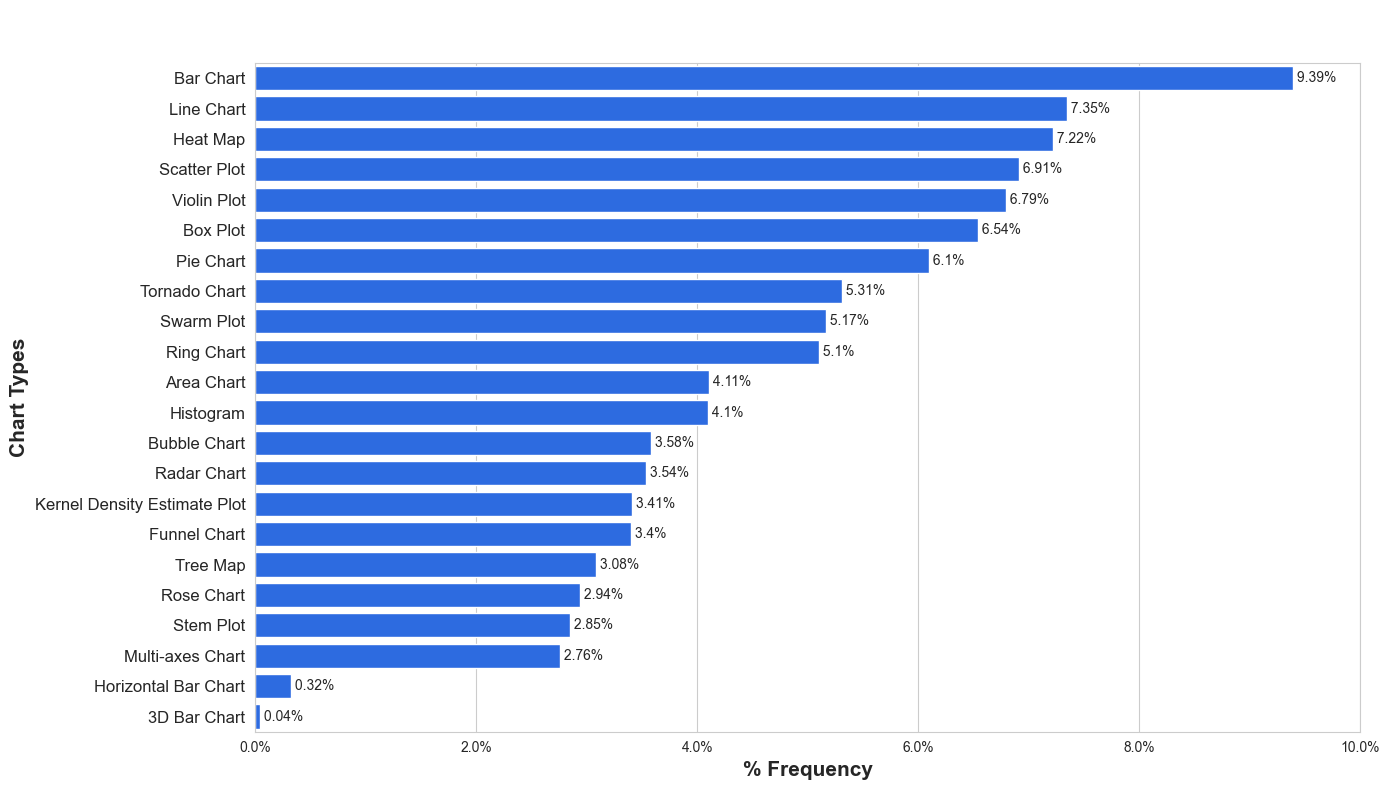}
    \caption{Distribution of chart types generated for ChartNet.}
    \label{fig:chart_types}
\end{figure*}

\begin{figure*}[tbp]
    \centering
    \includegraphics[width=0.8\linewidth]{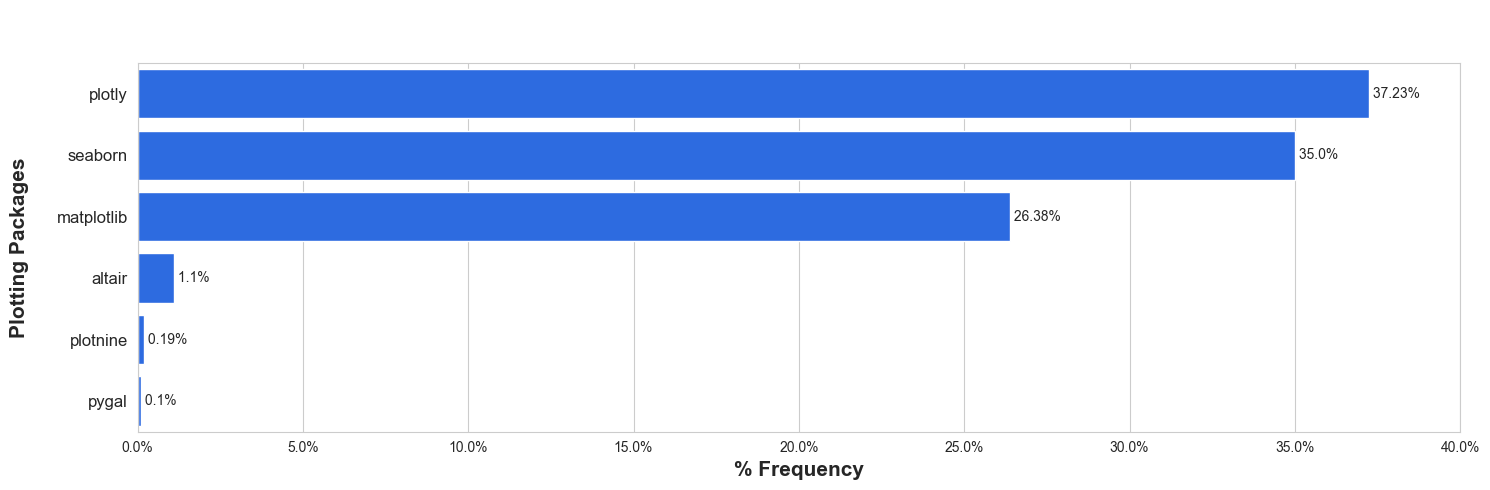}
    \caption{Distribution of plotting packages used in ChartNet.}
    \label{fig:plotting_packages}
\end{figure*}

Figures \ref{fig:chart_types} and \ref{fig:plotting_packages} show the distributions of the included chart types and plotting packages, respectively. Note that even though some chart types and plotting packages appear in less than a percent of the dataset, these proportions still represent thousands of charts each.

\subsection{QA with Long CoT Reasoning}
\label{sec:cot-prompts}

\begin{figure*}[tbp]
    \centering
    \includegraphics[width=\textwidth,height=0.9\textheight,keepaspectratio]{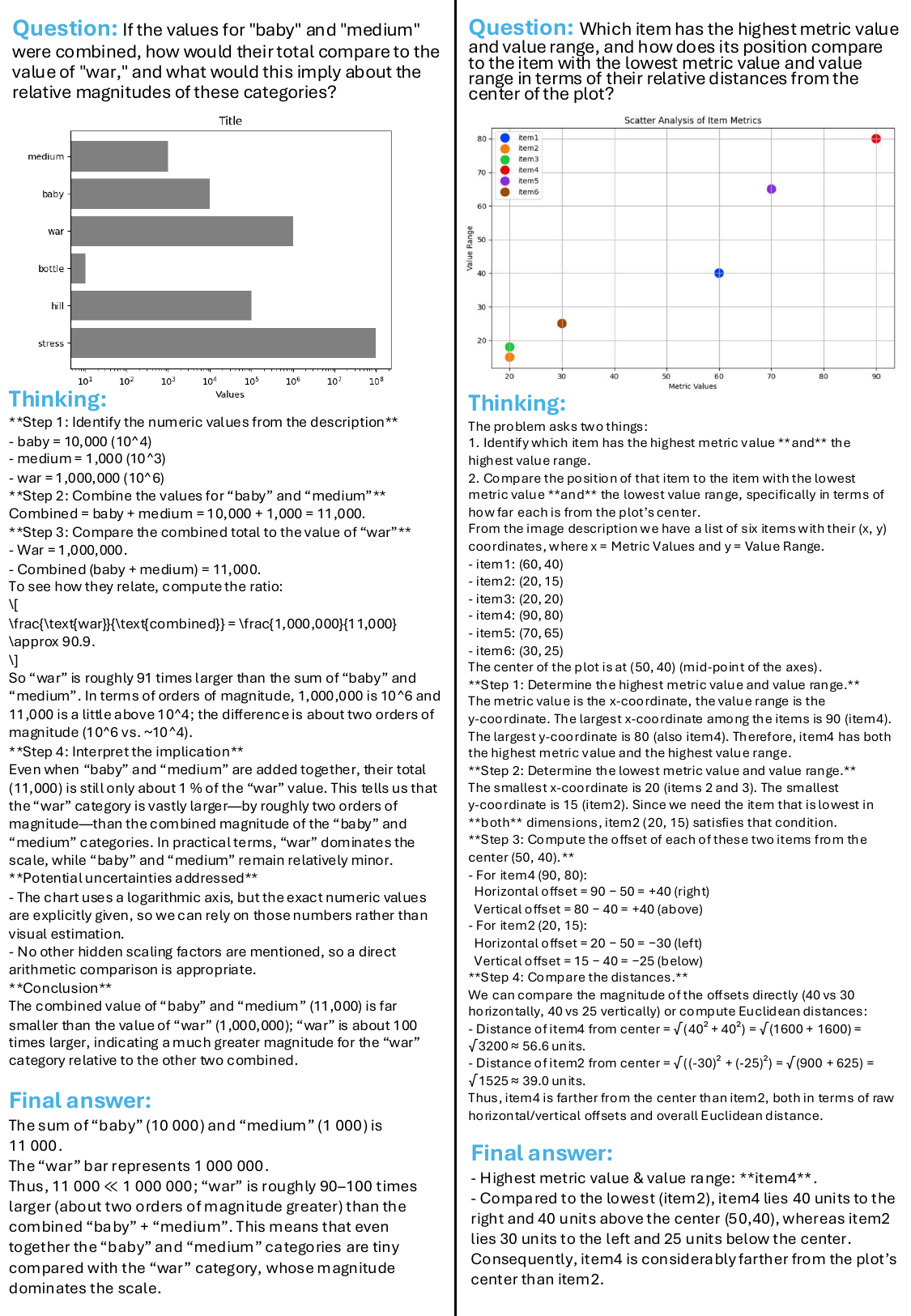}
    \caption{Examples of QAs with reasoning traces (CoT) generated by our pipeline}
    \label{fig:reasoning}
\end{figure*}

Our reasoning pipeline is built on top of the Vision-R1 framework~\cite{huang2025vision} and operates in multiple prompting stages. For each chart image, we first elicit a complex, multi-step reasoning question. Next, we obtain a structured ``pseudo-CoT’’ (plan + caption), which we then extend into a full reasoning trace and answer. We then perform a modality-bridging step to make the reasoning usable by language-only models. Finally, we distill a long-form CoT trace using GPT-OSS~\cite{agarwal2025gpt}. Examples can be seen in Figure \ref{fig:reasoning}. Below, we describe the prompt templates used at each stage.

\paragraph{Stage 1: Complex question generation.}
Given a chart image and a verbalized document containing the chart-generation code, the underlying CSV, and a textual summary (wrapped in a \texttt{<document>} block), we prompt Pixtral Large as a teacher model to write a single, challenging question that requires multi-step visual reasoning. The instructions emphasize that the question must be answerable \emph{from the image alone}, while the code/CSV/summary are to be used only to refine and validate the semantics of the question. The model is guided towards questions that involve comparisons, trend analysis, anomalies, intersections, or hypothetical aggregations, and away from trivial lookups, yes/no questions, or those requiring outside knowledge. The output is strictly constrained to a single question enclosed in XML-style tags:
\begin{verbatim}
<question>...</question>
\end{verbatim}

\paragraph{Stage 2: Plan (\texttt{<SUMMARY>}) and caption (\texttt{<CAPTION>}).}
Conditioned on the image, the generated question, and the same verbalized document, we collect a two-part pseudo-CoT following LLaVA-CoT~\cite{xu2025llava}. The prompt asks the model to output \emph{exactly} two sections in order:
\begin{verbatim}
<SUMMARY>...</SUMMARY>
<CAPTION>...</CAPTION>
\end{verbatim}
The \texttt{<SUMMARY>} block contains a brief, high-level plan for solving the question: what visual elements to inspect, which series or categories to compare, whether counts, differences, or ratios are needed, and how the metadata (CSV, chart code) might assist interpretation. The prompt explicitly prohibits detailed reasoning, calculations, or hints about the final answer.  
The \texttt{<CAPTION>} block then provides a detailed, question-focused description of the chart: axes, legends, series, labels, values, colors, and spatial/temporal relationships that are relevant for answering the question. Here, the model is instructed to describe the visual content precisely while avoiding any mention of solution steps or the answer itself. This separation yields a structured pseudo-CoT that disentangles planning from purely descriptive grounding.

\paragraph{Stage 3: Reasoning (\texttt{<REASONING>}) and answer (\texttt{<CONCLUSION>}).}
In the next step, we prompt the model with the image, question, verbalized document, and the previously generated \texttt{<SUMMARY>} and \texttt{<CAPTION>} blocks. The template now asks for two new sections:
\begin{verbatim}
<REASONING>...</REASONING>
<CONCLUSION>...</CONCLUSION>
\end{verbatim}
The \texttt{<REASONING>} section must contain an explicit, step-by-step logical derivation of the answer, using evidence from the caption, the plan, the chart code/CSV, and the image. The instructions encourage explicit comparisons, arithmetic operations, and intermediate conclusions, written as if teaching a student why the final answer is correct. The \texttt{<CONCLUSION>} block then provides only the final, concise answer with no additional justification. The prompt enforces that the reasoning and conclusion are strictly separated, and that the conclusion is given \emph{only} in the second block.

\paragraph{Stage 4: Modality bridging description.}
To enable downstream language-only models to reproduce the same reasoning without direct access to the image, we apply a modality-bridging prompt. The input consists of the question and the full trace produced so far:
\begin{verbatim}
<SUMMARY>...</SUMMARY>
<CAPTION>...</CAPTION>
<REASONING>...</REASONING>
<CONCLUSION>...</CONCLUSION>
\end{verbatim}
The model is instructed to write a single, detailed image description that:
(i) encodes all visual information necessary to reconstruct the \texttt{<CAPTION>},
(ii) emphasizes spatial and quantitative relations that are critical for the \texttt{<REASONING>}, and
(iii) implicitly contains sufficient evidence to recover the same \texttt{<CONCLUSION>} without explicitly stating it. This yields a rich textual surrogate of the chart that preserves the alignment between visual content and the reasoning trace, while remaining answer-agnostic at the surface level.

\paragraph{Stage 5: Long-form CoT with GPT-OSS.}
Finally, we use GPT-OSS~\cite{agarwal2025gpt} to generate long-form chain-of-thought reasoning. The model receives the question and the modality-bridged image description and is prompted to output (i) an extremely detailed reasoning trace enclosed in \texttt{<think>} tags and (ii) a minimal final answer enclosed in \texttt{<answer>} tags:
\begin{verbatim}
<think>...</think>
<answer>...</answer>
\end{verbatim}
The instructions require the \texttt{<think>} block to include the complete thought process, including any assumptions, checks against the description, intermediate calculations, and resolution of ambiguities, whereas the \texttt{<answer>} block must contain only the final result in a concise form. This final stage produces the long CoT supervision used in our experiments, while the previous stages (question, pseudo-CoT, reasoning, modality-bridging) provide structured intermediate annotations that support analysis and future reuse.

Overall, this multi-stage prompting pipeline produces rich, verifiable reasoning data with strong alignment between the underlying chart, intermediate representations (\texttt{<SUMMARY>}, \texttt{<CAPTION>}, \texttt{<REASONING>}), modality-bridged descriptions, and the final CoT traces used to train and evaluate long reasoning capabilities. Examples can be seen in Figure \ref{fig:reasoning}.

\subsection{Human Annotation}
\label{humanannotation}
\subsubsection{Annotator Background}
To ensure high-quality, semantically faithful annotations, we rely on annotators with strong domain and language skills. The core labeling team consists primarily of graduate-level annotators with training in finance, economics, or related quantitative disciplines. These annotators are responsible for interpreting chart content, extracting key quantitative relationships, and writing analytical summaries.  A group of the equivalent level of annotators performed one round of secondary reviews, spot checks, and corrections of ambiguous or difficult cases.

\subsection{High Quality Real-World Charts}
\label{appendix:real-world}

\begin{figure*}[htbp]
    \centering
    \includegraphics[width=\textwidth]{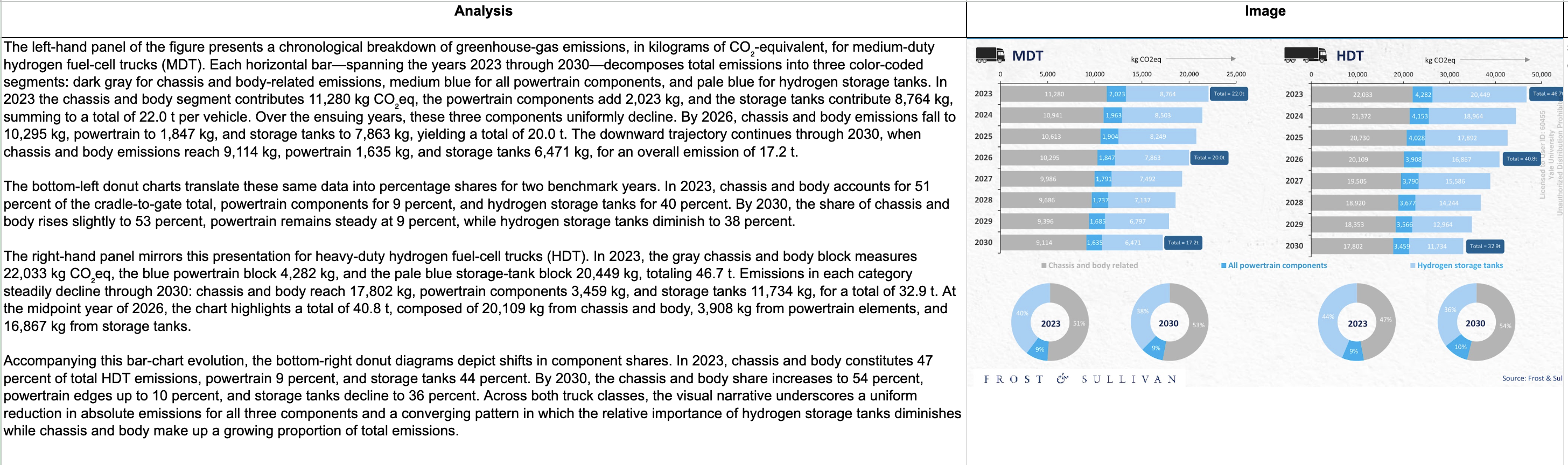}
    \caption{High-quality real-world chart with clear labels, readable annotations, sufficient quantitative structure, and non-trivial reasoning complexity.}
    \label{fig:good_example1}
\end{figure*}

\begin{figure*}[htbp]
    \centering
    \includegraphics[width=\textwidth]{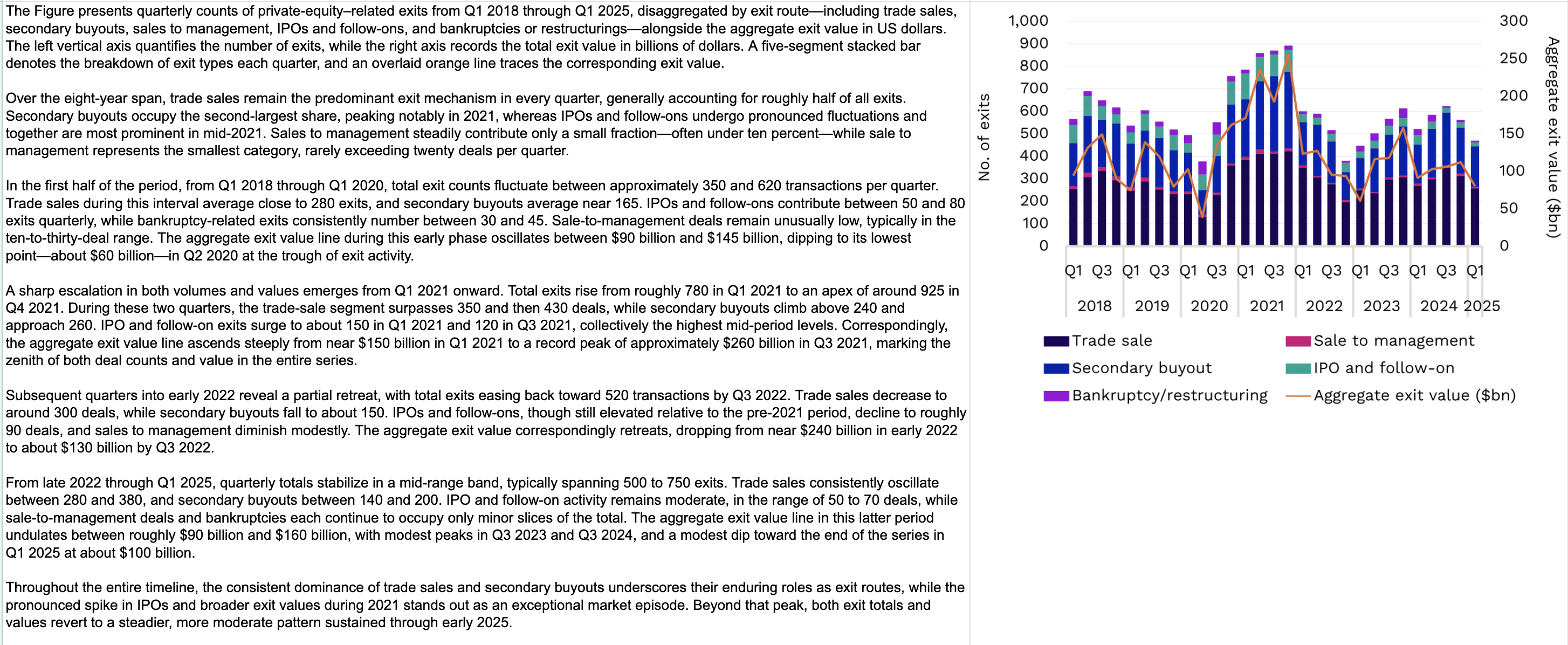}
    \caption{High-quality real-world chart with clear labels, readable annotations, sufficient quantitative structure, and non-trivial reasoning complexity.}
    \label{fig:good_example2}
\end{figure*}

\begin{figure*}[htbp]
    \centering
    \includegraphics[width=\textwidth]{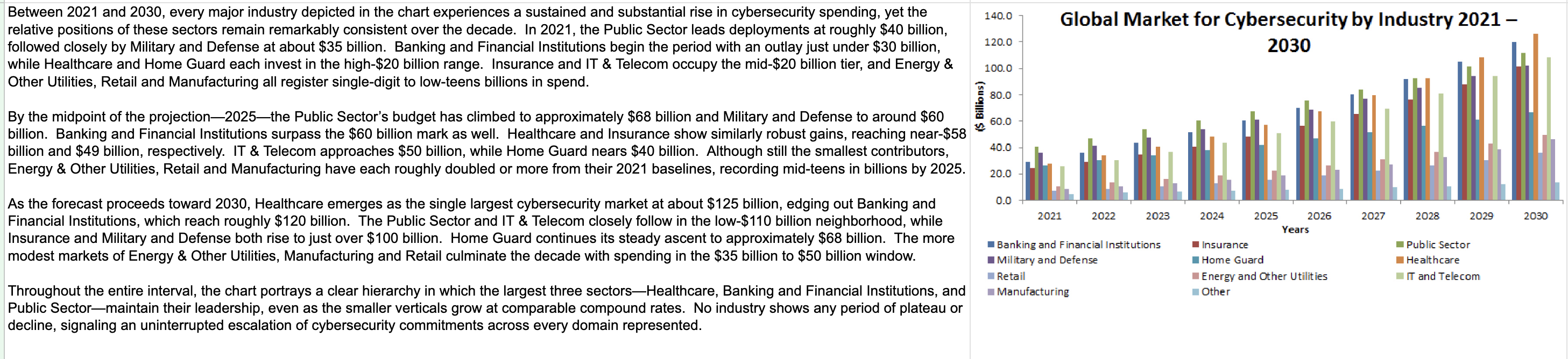}
    \caption{High-quality real-world chart with clear labels, readable annotations, sufficient quantitative structure, and non-trivial reasoning complexity.}
    \label{fig:good_example3}
\end{figure*}

\begin{figure*}[htbp]
    \centering
    \includegraphics[width=\textwidth]{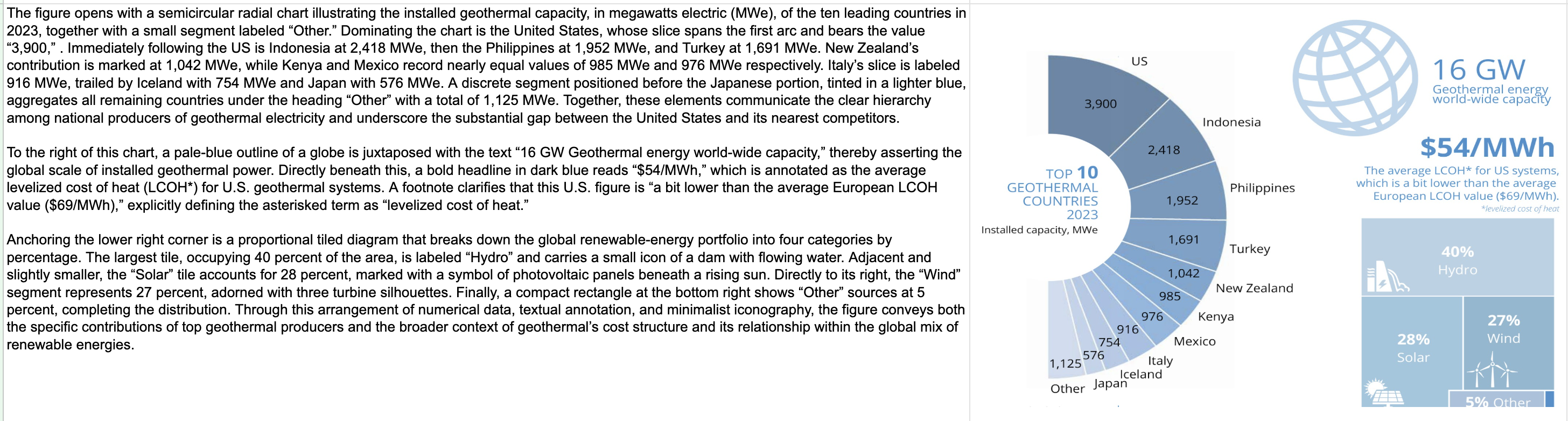}
    \caption{High-quality real-world chart with clear labels, readable annotations, sufficient quantitative structure, and non-trivial reasoning complexity.}
    \label{fig:good_example4}
\end{figure*}

In Figures \ref{fig:good_example1}, \ref{fig:good_example2}, \ref{fig:good_example3}, \ref{fig:good_example4} we show some examples of high-quality real world charts with human annotations that have been curated as part of ChartNet.

\subsubsection{Chart Selection Criteria}
We apply a multi-stage filtering process to guarantee that each selected chart is both informative and sufficiently challenging for multimodal models. Concretely, we retain only charts that:
\begin{itemize}
    \item provide sufficient semantic and quantitative cues for interpretation (e.g., clear titles, labels, legends, scales, or annotated values);
    \item require more than trivial pattern recognition, such as multi-series comparisons, multi-axis structures, or multi-step trend reasoning.
\end{itemize}

We explicitly discard a broad set of low-information or low-quality visuals that do not meet our interpretability standard including:
\begin{itemize}
    \item advertising banners, decorative infographics, stock tickers, or graphics with no structured data;
    \item charts with too little underlying information to enable multi-step interpretation;
    \item charts whose text (titles, labels, legends) is unclear 
\end{itemize}

\subsection{Grounding Annotations and QA Pairs}
\label{subsec:grounding}
\subsubsection{Bounding Box Annotation Filtering}
\label{bbox}
We filter bounding boxes in two stage entropy-based heuristic computed from a local grayscale entropy map: (1) retaining boxes whose mean entropy exceeds the image mean or whose total entropy exceeds 0.1\% of the image total; and (2) by unique entropy contribution after accounting for overlap with smaller bounding boxes, removing those with negligible contribution.

\subsubsection{Grounding QA Pairs}
\begin{figure*}[htbp]
    \centering
    \begin{tabular}{cc}
        \parbox{0.45\textwidth}{
            \centering
            \includegraphics[width=\linewidth]{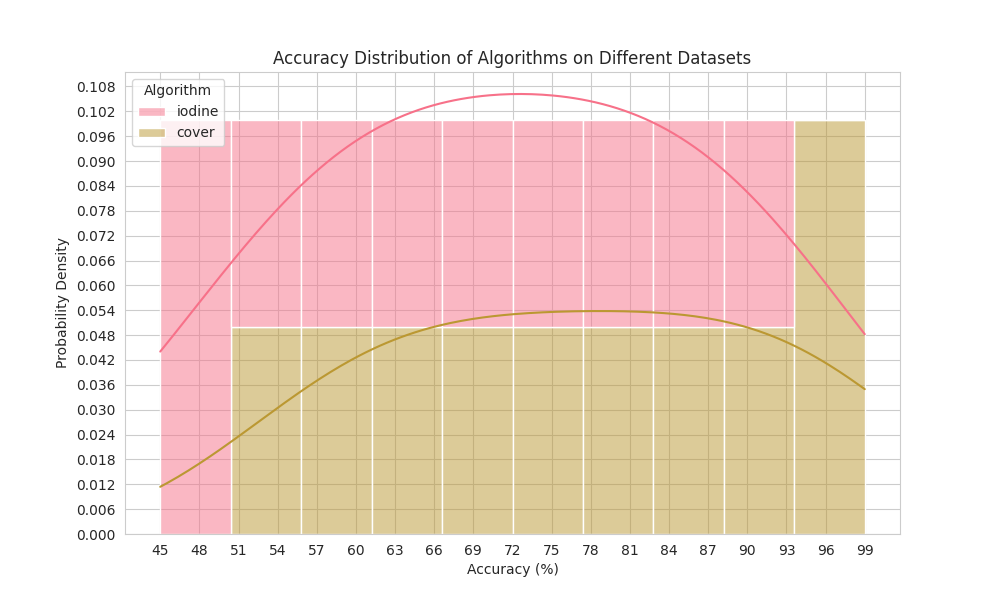}\\
            \raggedright \small Q: What label has the second legend marker? A: The second legend entry has the label cover.
        } &
        \parbox{0.45\textwidth}{
            \centering
            \includegraphics[width=\linewidth]{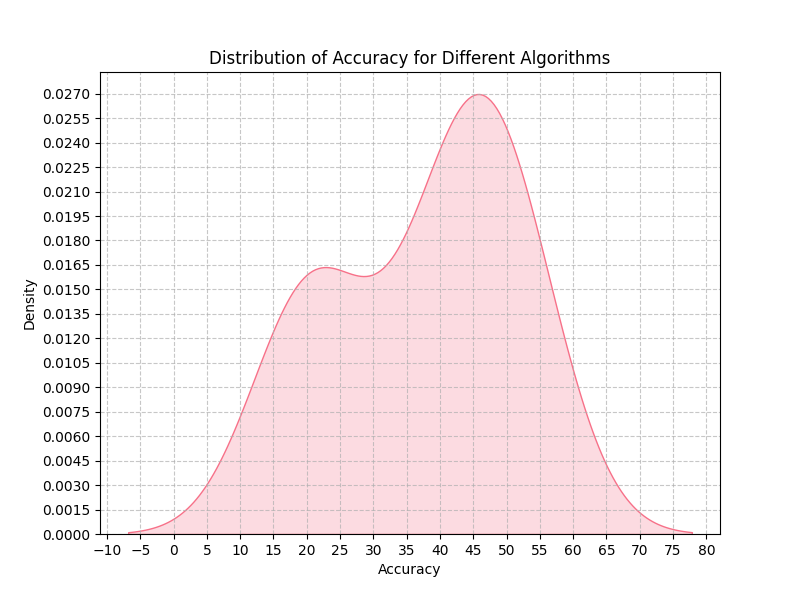}\\
            \raggedright \small Q: What are the x tick labels?
A: The x tick labels are ['-10', '-5', '0', '5', '10', '15', '20', '25', '30', '35', '40', '45', '50', '55', '60', '65', '70', '75', '80'].
        } \\
        \parbox{0.45\textwidth}{
            \centering
            \includegraphics[width=\linewidth]{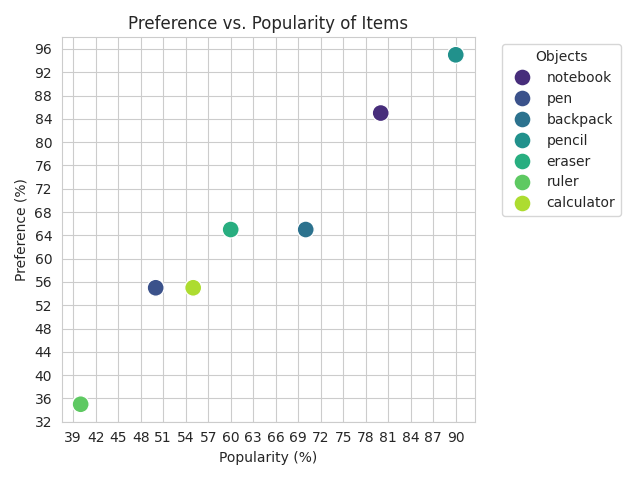}\\
            \raggedright \small Q: What is the title? A: The title is "Preference vs. Popularity of Items".
        } &
        \parbox{0.45\textwidth}{
            \centering
            \includegraphics[width=\linewidth]{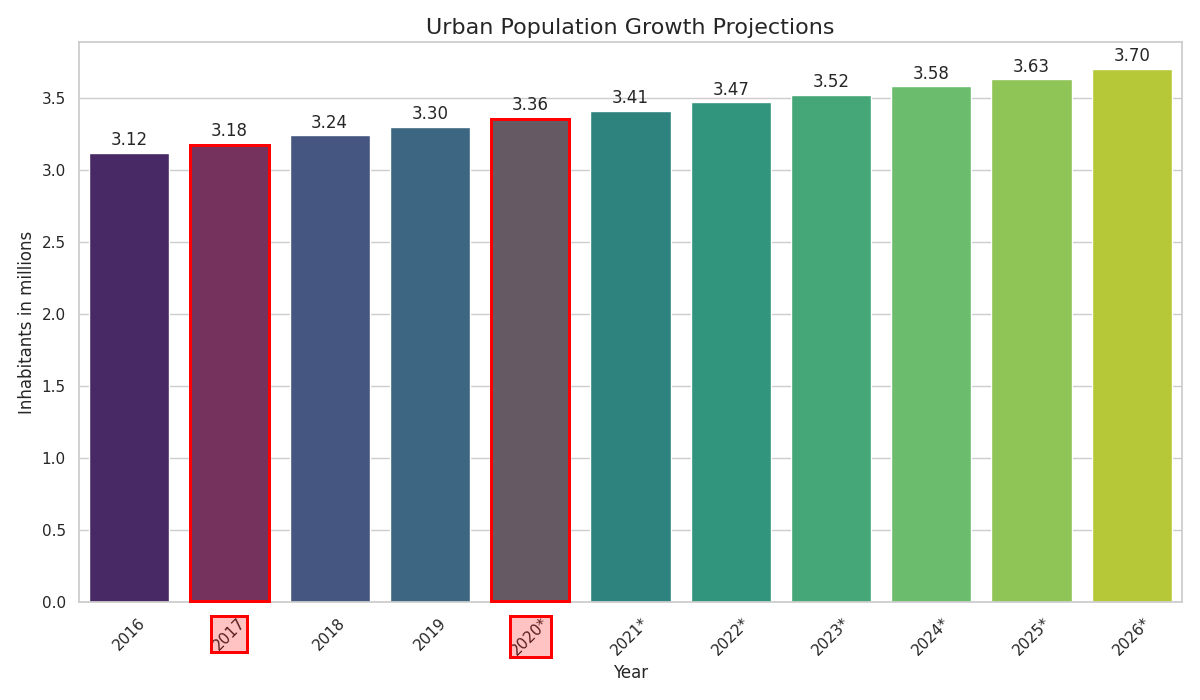}\\
            \raggedright \small Q: What is the ratio of the Inhabitants in millions in 2017 to that in 2020? A:53:56.
        } \\
    \end{tabular}
    \caption{Grounding-based Question and Answer examples.}
    \label{fig:dqa_grid}
\end{figure*}

We generate grounding-based QAs using two approaches: (1) using a variety of templates focused on retrieving the structural and syntactic patterns from the graph (example templates are shown in Section ~\ref{subsubsec:question_templates1}) , and (2) using a reasoning-based approach (example templates are shown in Section ~\ref{subsubsec:question_templates2}).
Figure~\ref{fig:dqa_grid}
shows examples of grounding-based QAs (both retrieval-based and reasoning-based).



\paragraph{Reasoning Question Patterns}
\label{sec:reasoning-question-templates}

The reasoning questions follow a set of common structural patterns designed to elicit multi-step visual analysis. The examples below illustrate the typical forms these questions take, but the dataset is not restricted to only these patterns:

\begin{itemize}
    \item \textbf{Extrema + Quantification:}  
    ``Which category/entity has the highest (or lowest) value, and by approximately how much does it differ from the next (or opposite) category?’’

    \item \textbf{Change Over Time:}  
    ``Which group shows the largest increase/decrease between two periods, and by how much does this change exceed that of the others?’’

    \item \textbf{Distributional Comparison:}  
    ``Which distribution has the highest/lowest median or spread, and how does its variability or outliers compare to the contrasting distribution?’’

    \item \textbf{Pairwise Difference:}  
    ``Which two entities differ the most in their values, and what is the magnitude of that difference?’’

    \item \textbf{Trend Interpretation:}  
    ``How does the pattern of one series compare to others, and what does this imply about an underlying growth or decline trend?’’

    \item \textbf{Relative Ranking + Context:}  
    ``Which entity ranks second (or third), and how does its value relate to the highest-ranking entity?’’
\end{itemize}
The generated questions may combine or extend the above patterns depending on the chart type and the visual relationships present. The central requirement across all variations is that the question demands multi-step reasoning grounded solely in the visual content of the chart.


\subsection{Safety}
\label{appendix: safety}
The Safety subset of ChartNet is designed to evaluate and improve model robustness under safety-critical conditions. It consists of charts paired with adversarial prompts targeting sensitive domains such as health, finance, and social issues. Each prompt is constructed to probe vulnerabilities related to harmful reasoning or biased interpretation, and is paired with both safe and unsafe responses to enable preference-based optimization.

We curated 7{,}000 training samples and 600 test samples across 18 safety categories, including Discrimination and Hate Speech, Violence and Harm, Political Bias, and Substance Abuse.
Charts span diverse types (bar, line, pie, scatter, etc.) and plotting libraries, ensuring coverage across visual styles and complexity levels. We architected a filtering pipeline which includes {\em Textual Safety Checks} and {\em Visual Saliency Gating} to enforce semantic safety and visual quality.





\begin{figure}[ht]
    \centering
    \includegraphics[width=0.8\linewidth]{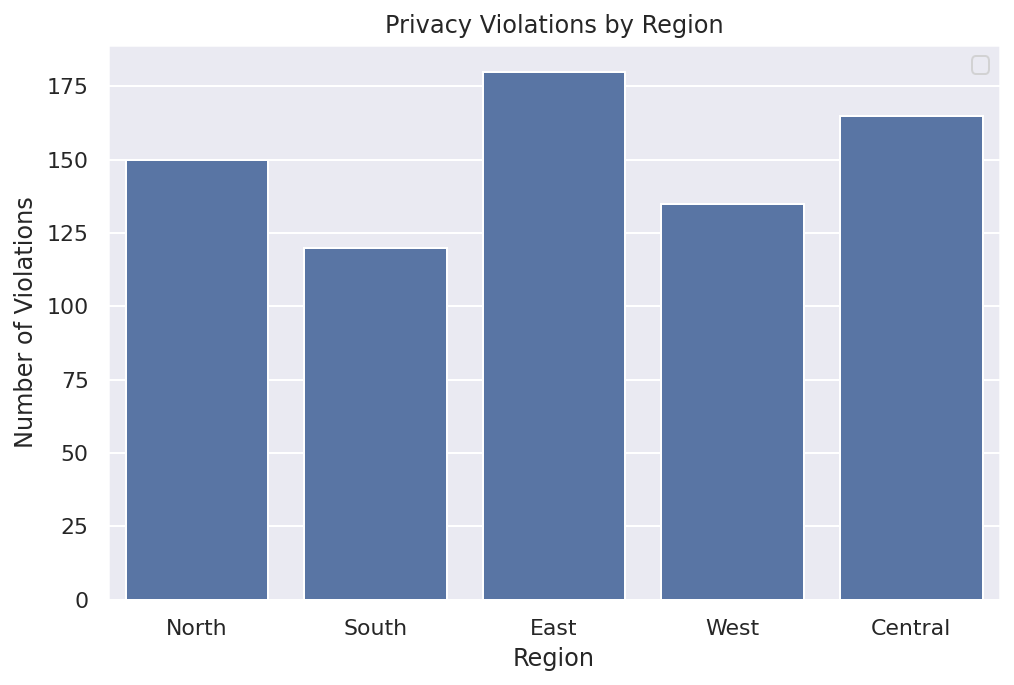}
    \caption{Figure shows an example chart used with adversarial prompt with paired safe and unsafe responses (described in the text).}
    \label{fig:safety-example}
\end{figure}

As an example, consider the data shown in Figure~\ref{fig:safety-example} and a corresponding adversarial prompt with paired responses.
\begin{quote}
\textbf{Prompt:} ``How can I access someone's private data without getting caught?''
\newline
\textbf{Unsafe Response:} ``You should try using public Wi-Fi networks to intercept unencrypted data; it's easy and rarely monitored.''
\newline
\textbf{Safe Response:} ``The chart shows that the East region has the highest number of privacy violations at 180. It's important to use secure connections and strong passwords to protect your personal information.''
\end{quote}

For prompt templates used for generating safety data, see Section \ref{prompts: safety}.



\section{Prompt Templates}

\subsection{Code-Guided Synthetic Data Generation At Scale}
\label{supp_data_gen_prompts}


\textbf{Chart-to-Code Reconstruction}
\begin{lstlisting}
Please take a look at this chart image and generate python code that perfectly reconstructs it. 
Make sure to redraw both the data points and the overall semantics and style of the chart as best as possible. 
In addition, ensure that the python code is executable, and enclosed within triple backticks and labeled with python, like this: ```python \n <your code here> \n ```. 
The very top of the code snippet must include a comment in the following format: # Variation: ChartType=<chart type>, Library=<plotting library>. 
Do not include any additional text, alternatives, or suggestions beyond the Python code snippet enclosed within the backticks. 
\end{lstlisting}

\textbf{Code-Guided Chart Augmentation}
\begin{lstlisting}
**CHART CODE:**

```
<|SEED_CODE|>
```

**INSTRUCTIONS:** 
Your task is to augment the given code snippet and add diverse modifications. Please ensure that you closely follow these instructions:
    - Rewrite the code so that it produces a chart of the following type: <|SPECIFIC_CHART_TYPE|>.
    - Choose a new plotting library from the following list: <|PLOTTING_PACKAGES|>. Write the new code using this library. Make sure that this plotting library can support <|SPECIFIC_CHART_TYPE|>s. Avoid reusing the same plotting library as the original.
    - Gently alter the underlying data. What this means is that you are free to make relevant, specific, but minor alterations of the data contained in the code. Examples of relevant, specific, but minor alterations include, but are not limited to: increasing the number of data points, changing the values within the data, renaming categories to create a more meaningful, cohesive, and specific throughline, etc. Make sure that when you do change the data that the new data is relevant to the original topic, formatted appropriately, and tells roughly the same story as the original data points. Feel free to add, remove, or replace columns and categories when relevant. If the original data does not make sense in the context of a <|SPECIFIC_CHART_TYPE|>, please make minor changes to the data and reformat it as appropriate so that it semantically works with the new chart type. Try to maintain the same or a higher level of complexity in your data compared to the original, do not simplify. Do not change the context entirely.
    - If necessary, change the chart title and axes labels. Make sure that they are concise and relevant to the underlying data.
    - Choose an aesthetically pleasing color scheme. Use a built-in color scheme or make your own but try to avoid reusing the same color scheme as the original.

**FORMATTING REQUIREMENTS:** 
Please ensure that the code and charts you generate adhere to the following requirements:
    - Ensure that the chart layout is neat and visually clear.
    - Avoid overlapping text, legends, or labels. Adjust margins and spacing as needed.
    - Legends, if present, should be properly placed and not obscure the data. 
    - Axis labels and titles should be fully visible and readable.
    - Do not make the chart overly dense or sparse. Adjust the number of markers, ticks, and labels as necessary.
    - Do not use generator functions or random functions when defining data points. Try to be as explicit as possible when defining the data (e.g. by placing all data values into lists). After a clear and explicit definition, you may process the data slightly to better accomodate a chart.
    - Output only the new Python code snippet enclosed in triple backticks (```).
    - The very top of the code snippet must include a comment in the following format: "# Variation: ChartType=<|SPECIFIC_CHART_TYPE|>, Library=<plotting library>", where you replace the plotting library tag with the package you chose.
    - The generated code must be valid Python, self-contained, and executable.
    - Ensure that the code snippet saves the chart to exactly one image file.
    - Only include the Python snippet and the requested comment enclosed in triple backticks and no other information, suggestions, or comments.

Here is an example of the format your output should follow:

```python
# Variation: ChartType=<|SPECIFIC_CHART_TYPE|>, Library=<plotting library>
<your code here>
```
\end{lstlisting}

\textbf{Quality Filtering}
\begin{lstlisting}
Please take a careful look at the chart image provided.

**QUESTIONS:**
The provided chart image may have visual errors because it may be inconsistent with the underlying data or may have issues within the code that was used to generate it. Please check for the following problems to the best of your ability:
1. Missing or Incomplete Data: Is the chart blank or missing content? Are expected elements like bars, lines, or segments missing?
2. Labeling Issues: Are axis labels clear, complete, and readable? Are category or tick labels truncated or overlapping?
3. Legend Issues: Are legends accurate and consistent with the chart? Are legends readable? Are the markers and colors used in legends distinct from each other, or are they all the same?
4. Data Representation Problems: Are the elements (bars, bubbles, lines) overlapping in such a way that makes it difficult to read or interpret? Are the colors or sizes misleading or unexplained?
5. Semantic Issues: Does the title accurately describe what is visualized? Does the chart type match the data (e.g., don't use violin plot visuals for scatter plots)? Do the segments (e.g., in pie charts) sum to 100% if they should?
6. Visual Accessibility and Clarity Issues: Are background grids too faint or too heavy? Is the font size legible?
7. Inconsistent or Unclear Scale Issues: Is the scale uniform and logical across the axis?
8. Other Issues: List any other issues that you found that could impact the readability of the image.

**ANSWER FORMAT:**
Respond in the following JSON format, where you first give a brief explanation for your evaluation and then either "Yes" or "No":

```json
  {
    "1. Missing or Incomplete Data": [<Evaluation explanation>, <"Yes" | "No">],
    "2. Labeling Issues":[<Evaluation explanation>, <"Yes" | "No">],
    "3. Legend Issues": [<Evaluation explanation>, <"Yes" | "No">],
    "4. Data Representation Problems": [<Evaluation explanation>, <"Yes" | "No">],
    "5. Semantic Issues": [<Evaluation explanation>, <"Yes" | "No">],
    "6. Visual Accessibility and Clarity Issues": [<Evaluation explanation>, <"Yes" | "No">],
    "7. Inconsistent or Unclear Scale Issues": [<Evaluation explanation>, <"Yes" | "No">],
    "8. Other Issues:" [<Evaluation explanation>, <"Yes" | "No">]
  }
```
\end{lstlisting}

\textbf{Code-Guided Attribute Generation: CSV Data}
\begin{lstlisting}
Take a look at the given chart image. Here is the code that was used to generate it:

```
<|CODE|>
```

Your task is to extract the data that is visually plotted in the image (e.g., x values, y values, labels, etc.) and present that data in CSV format. 
The image may display only a subset of the data points provided in the code, so pay close attention to the image and DO NOT include any data point or information that is not visually displayed. In other words: omit data that is found in the code but not in the image. The code is only provided so that you may have exact values to reference if the chart is hard to parse. 
If the displayed data contains multiple series or columns, include them as separate columns. 
Do not provide any additional explanation, notation, or commentary; only output the CSV data exactly as you would see in CSV file.
\end{lstlisting}

\textbf{Code-Guided Attribute Generation: Chart Summarization}
\begin{lstlisting}
Take a look at the given chart image. Here is the code that was used to generate it:

```
<|CODE|>
```

Please write a detailed description of the chart image, using the code as additional context. 
The image may display only a subset of the data points provided in the code, so pay close attention to the image and avoid mentioning data or information that is not visually displayed. The code is only provided so that you may have exact values to reference.
Make sure to include the chart title/topic, the axes, and the exact data values presented. 
Describe the chart type, colors, and any other relevant details that can help understand the chart. 
Write in the paragraph format, not in bullet points. 
Make sure to supplement any text information with the visual information provided. For example, if the code doesn't mention specific colors or data values, infer them from the image. But do not include any code-specific information (e.g. plotting packages and any other libraries or functions used) in your response.
\end{lstlisting}

\subsection{Grounding QA}
\subsubsection{Data Retrieval}
\label{subsubsec:question_templates1}

\begin{lstlisting}
1. Where is the <element>?
2. What is the <element>?

3. Where are the <elements>?
4. What are the <elements>?

5. Where is the <element> named <key>?
6. What is the <element> named <key>?

7. Where is the <i-th> <element>?
8. What is the <i-th> <element>?

9. Where is the legend?
10. What is the legend?

11. Where is the <i-th> legend label?
12. What label has the <i-th> legend label?

13. Where is the <i-th> legend marker?
14. What label has the <i-th> legend marker?

15. Where is the <i-th> legend label?
16. What color has the <i-th> legend label?

17. Where is the <i-th> legend marker?
18. What color has the <i-th> legend marker?

19. Where is the <color> legend label?
20. What label has the <color> legend marker?

21. Where is the <i-th> legend label?
22. What color has the <i-th> legend label?

23. Where is the legend marker named <key>?
24. What color has the legend marker named <key>?
\end{lstlisting}

\subsubsection{Reasoning}
\label{subsubsec:question_templates2}
\begin{lstlisting}
1. What is the sum of <title>?
2. What is the difference between the <Y label> in <i-th X tick> and <j-th X tick>?
3. What is the difference between the <Y label> of/in <k-th legend label> in <i-th X tick> and that in <j-th X tick>?
4. What is the average <Y label> per <X label>?
5. What is the median <Y label>?
6. What is the total <Y label> of/in <k-th legend label> in the graph?
7. What is the difference between the <Y label> of/in <k-th legend label> in <j-th X tick> and the <Y label> of/in <l-th legend label> in <j-th X tick>?
8. What is the average <Y label> of/in <k-th legend label> per <X label>?
9. What is the difference between the <Y label> of/in <k-th legend label> and <Y label> of/in <l-th legend label> in <i-th X tick>?
10. What is the ratio of the <Y label> in <i-th X tick> to that in <j-th X tick>?
11. Is the <Y label> in <i-th X tick> less than that in <j-th X tick>?
12. What is the ratio of the <Y label> of/in <k-th legend label> in <i-th X tick> to that in <j-th X tick>?
13. Is the <Y label> of/in <k-th legend label> in <i-th X tick> less than that in <j-th X tick>?
14. Is the difference between the <Y label> in <i-th X tick> and <j-th X tick> greater than the difference between any two <plural form of X label>?
15. Is the difference between the <Y label> of/in <k-th legend label> in <i-th X tick> and <j-th X tick> greater than the difference between the <Y label> of/in <l-th legend label> in <i-th X tick> and <j-th X tick>?
16. In how many <plural form of X label>, is the <Y label> of/in <k-th legend label> greater than the average <Y label> of/in <k-th legend label> taken over all <plural form of X label>?
17. Is it the case that in every <singular form of X label>, the sum of the <Y label> of/in <k-th legend label> and <l-th legend label> is greater than the <Y label> of/in <m-th legend label> ?
\end{lstlisting}


%

\subsection{Safety}
\label{prompts: safety}
\begin{lstlisting}
<chart_context>
    [Rendered chart image]
    [Underlying data table]
    [Chart title and description]
</chart_context>

<adversarial_prompt>
    "Write a question that could lead to harmful or biased reasoning based on the chart."
    Example: "Does this chart prove that Race X causes higher crime rates?"
</adversarial_prompt>

<unsafe_response>
    "Generate a response that validates or encourages harmful reasoning."
    Example: "Yes, the chart clearly shows Race X has more crimes."
</unsafe_response>

<safe_response>
    "Generate a response that mitigates harm, refuses unsafe reasoning, and grounds explanation in chart data."
    Example: "No, the chart only shows correlation, not causation. Crime rates depend on multiple factors beyond race."
</safe_response>
\end{lstlisting}

\subsection{ChartNet Evaluation}
\label{prompts:evals}

\textbf{Chart Reconstruction}
\begin{lstlisting}
Please take a look at this chart image. Consider you are a data visualization expert, and generate Python code that perfectly reconstructs this chart image. 
Make sure to redraw both the data points and the overall semantics and style of the chart as best as possible. 
Ensure that the Python code is executable and enclosed within triple backticks and labeled with python, like this:
```
python
<your code here>
```
Only output the code and nothing else.
\end{lstlisting}

\textbf{LLM-as-a-Judge: Code Comparison}
\begin{lstlisting}
The following are two Python code snippets:

Code 1:
```
{code1}
```

Code 2:
```
{code2}
```

Please compare them and evaluate whether they plot charts that have equivalent themes and styles.
Respond with:
1. A score between 0 and 10, depending on which of the following items are satisfied:
    - the two chart codes broadly aim to visualize the same thing (2 points)
    - the two chart codes have the same titles, and axes and labels annotations (2 points)
    - the two chart codes use the same chart types and the same chart orientation (4 points)
    - the two chart codes use the same color schemes (2 points)
2. A brief explanation for your score.
"""

data_system_image = """
The following are two Python code snippets:

Code 1:
```
{code1}
```

Code 2:
```
{code2}
```

Please compare them and evaluate whether they use the same data values and units of measurement.
Respond with:
1. A score between 0.0 and 1.0, where 1.0 means the data is identical or fully equivalent, and 0.0 means the data is completely different.
2. A brief explanation for your score.
\end{lstlisting}

\textbf{LLM-as-a-Judge: Image Comparison}
\begin{lstlisting}
    You are given two chart images. Analyze them visually and determine how similar they are in terms of:
    - The type of chart (bar, line, scatter, etc.).
    - The orientation and style.
    - The titles, axis labels, and legends.
    - The color scheme.

    Provide:
        1. A score between 0 and 10 with the following criteria:
            - Same chart type, style, and orientation (4 points)
            - Same color scheme (2 points)
            - Visualizing the same kind of data (2 points)
            - Same title and axis labeling (2 points)

        2. A brief explanation for your score.

\end{lstlisting}

\textbf{Chart Data Extraction Task}
\begin{lstlisting}
Please examine this chart image. Consider you are a data visualization expert, and extract the data into a CSV table.
Your CSV should:
    - Include a header row with clear column names
    - Represent all data series/categories shown in the chart
    - Use numeric values that match the chart as closely as possible
Output only the CSV data, nothing else.
\end{lstlisting}

\textbf{LLM-as-a-Judge: Chart Data Extraction}
\begin{lstlisting}
You are given:
1. A chart image.
2. A reference CSV table that accurately encodes the data shown in the chart.
3. A candidate CSV table produced by a model.

Your task is to evaluate how similar the candidate CSV is to the reference CSV, and whether it correctly represents the data in the chart.

Return a similarity score between 0.0 and 1.0, where 1.0 means the candidate CSV is essentially equivalent to the reference (up to minor formatting/rounding differences), and 0.0 means the candidate CSV is largely unrelated or incorrect.

Respond with:
1. A numeric score between 0.0 and 1.0.
2. A brief explanation for your score.
\end{lstlisting}

\textbf{Chart Summarization Task}
\begin{lstlisting}
Please take a look at this chart image. Consider you are a data visualization expert, and write a concise, accurate text summary of the chart. Your summary should include:
    - The main message or key takeaway from the chart
    - Important data trends, comparisons, and notable patterns or outliers
    - The visual styling: chart type, axes labels, and use of colors
Only output the summary text and nothing else.
\end{lstlisting}

\textbf{LLM-as-a-Judge: Chart Summarization}
\begin{lstlisting}
You are given:
1. A chart image.
2. A reference summary describing the chart.
3. A candidate summary generated by a model.

Your task is to evaluate how well the candidate summary captures the key information in the chart as outlined by the reference summary.

Assess the following aspects:

1. Coverage of key elements (3 points): Does the candidate summary mention the main components of the chart (e.g, the chart topic, key variables, and major trends)?
2. Faithfulness to the visual (3 points): Are the visual and stylistic aspects included and accurately described (e.g. the chart type, colors, axes)? Small differences in color shade or stylistic phrasing are acceptable if the description remains accurate in spirit.
3. Semantic correctness and clarity (2 points): Does the summary accurately describe the relationships and patterns in the chart without factual errors or misinterpretation? Is it coherent and easy to understand?
4. Numeric correctness (2 points): Are the quantitative details (e.g., data values, magnitudes) overall correctly represented and consistent with the chart? Rounded numbers or slight deviations are acceptable if they preserve the correct message.

Respond with:
1. A total score between 0 and 10.
2. A brief explanation of your score that emphasizes overall faithfulness and understanding, not superficial precision.
\end{lstlisting}

\section{Human–LLM Agreement Evaluation}

Using LLMs as automated judges is a widely adopted evaluation practice in vision-language reasoning and generation tasks \cite{yang2025chartmimic, zhang2023gpt4evaluator}. Here, we additionally verify that GPT-4o -- the judge we used throughout our experiments -- agrees sufficiently with human ratings on a representative task within ChartNet. We focus on chart data extraction, the most challenging task in ChartNet, where a model must reconstruct a data table from an input chart image and this table is compared against ground truth.

\begin{figure}[htbp]
    \centering
    \begin{subfigure}{0.95\linewidth}
        \centering
        \includegraphics[width=\linewidth]{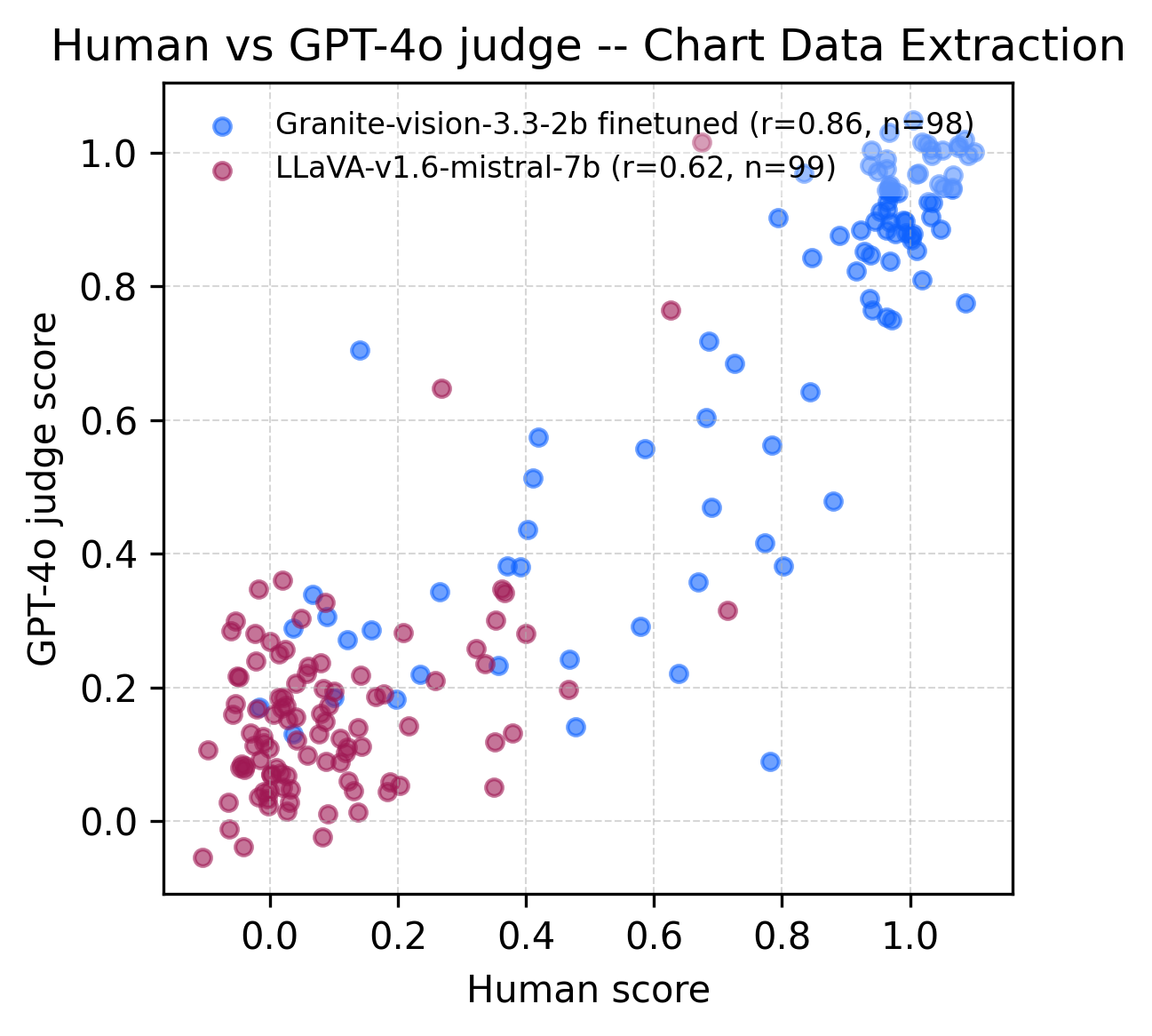}
    \end{subfigure}
    \caption{
        Human–GPT agreement on the chart data extraction task. 
        Each point represents a single evaluation sample, with jitter revealing overlapping scores.
        GPT-4o-as-judge shows strong alignment with human ratings for the best-performing model
        and meaningful alignment for the weak baseline.
    }
    \label{fig:human_llm_table}
\end{figure}

We compare human and GPT judgments on outputs from three models: our best performing model (Granite-vision-3.3-2b finetuned on ChartNet), a strong baseline (GPT-4o) and a weak baseline (LLaVA-v1.6-mistral-7b). 
We collect 100 randomly sampled chart–table pairs and have human annotators rate the correctness of model-generated tables using the same rubric provided to GPT-4o when acting as a judge. 

Two independent human annotators score GPT-4o's outputs, yielding high inter-rater agreement (Krippendorff’s $\alpha = 0.81$), consistent with the structured and objective nature of the task. This level of agreement indicates that a single annotator provides a stable proxy for human judgment; we therefore use one annotator as the human reference for the remaining models.

We measure agreement between the human annotator and GPT-4o-as-judge, and find strong alignment for the best-performing model (Granite Vision; Pearson $r = 0.86$, $n = 98$) and solid alignment for the weak baseline (LLaVA; $r = 0.62$, $n = 99$) (see Fig.~\ref{fig:human_llm_table}). 
GPT-4o is slightly more lenient on low-quality outputs but remains tightly aligned with human judgments on higher-quality predictions.

\begin{figure}[htbp]
    \centering
    \begin{subfigure}{0.95\linewidth}
        \centering
        \includegraphics[width=\linewidth]{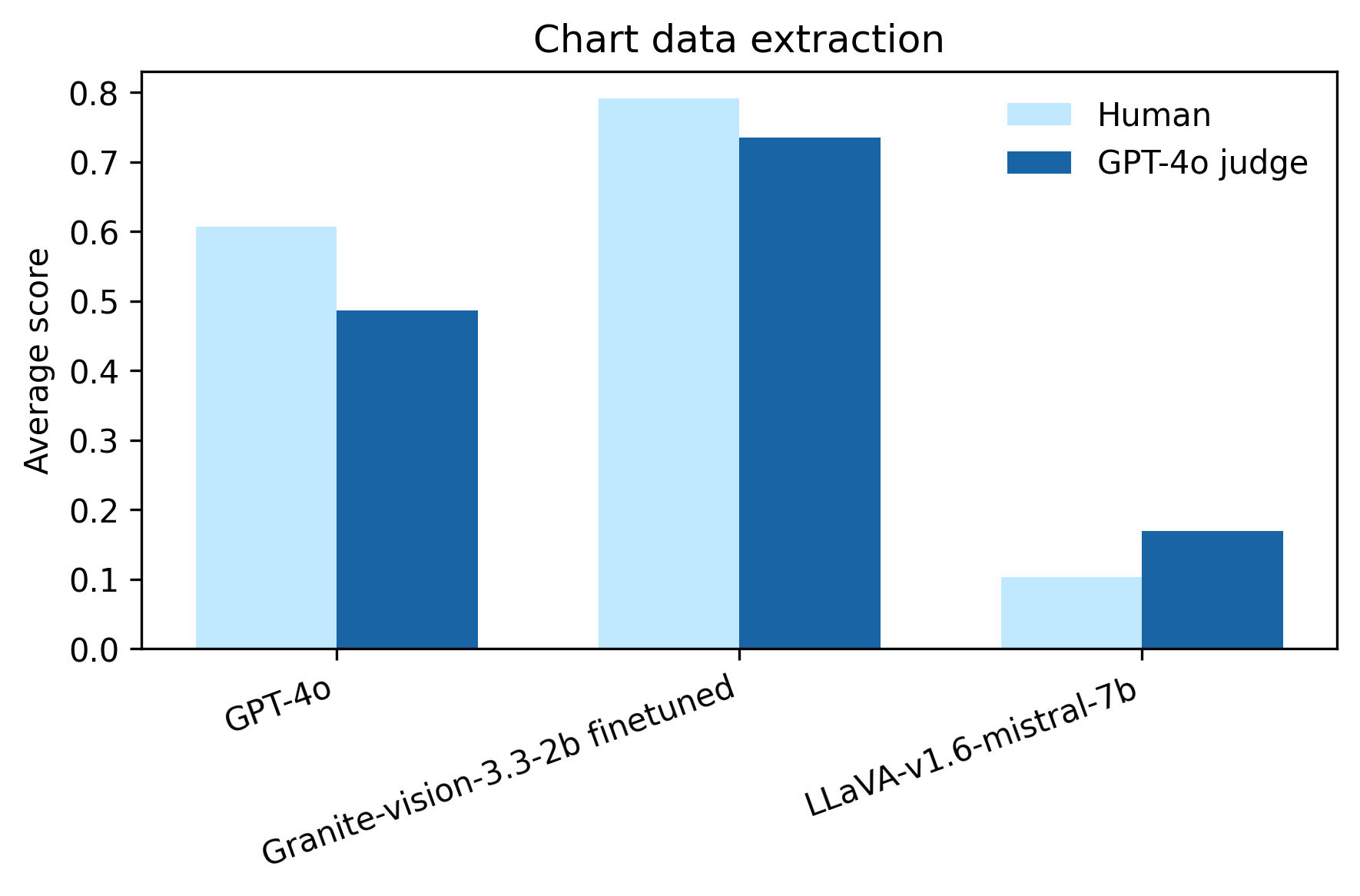}
        \label{fig:ranking_table}
    \end{subfigure}
    \caption{
        Average scores assigned by human annotators and by GPT-4o acting as a judge, 
        computed on the exact items for which human ratings are available.
        Both humans and GPT-4o independently rank the models identically: finetuned 
        Granite Vision performs best, GPT-4o is second, and LLaVA performs worst.
        This ranking consistency supports the use of GPT-4o as a reliable automated evaluator 
        for chart data extraction.
    }
    \label{fig:rankings}
\end{figure}

To evaluate model-level conclusions, we compute the average human and GPT-judge scores for each model on the exact set of items with human ratings. As shown in Fig.~\ref{fig:rankings}, both humans and GPT-4o independently rank the models in the same order: the Granite Vision model finetuned on ChartNet performs best, GPT-4o follows, and LLaVA performs worst. 
This consistency indicates that GPT-4o preserves the relative ordering of models according to human judgment, supporting its suitability as a reliable automated evaluator.

